\documentclass[a4paper,fleqn]{cas-dc}

\usepackage[authoryear]{natbib}
\usepackage{amsmath,amsfonts}
\usepackage{algorithmic}
\usepackage{algorithm}
\usepackage{array}
\usepackage{textcomp}
\usepackage{stfloats}
\usepackage{url}
\usepackage{autobreak}
\usepackage{verbatim}
\usepackage{graphicx}
\usepackage{verbatim}
\usepackage{tikz}
\usepackage{threeparttable}  
\usepackage{booktabs}
\usetikzlibrary{positioning}
\usetikzlibrary{shapes,arrows}
\usepackage{tabularx,ragged2e,siunitx}
\usepackage{multirow}	
\usepackage{color, colortbl}
\usepackage{hyperref}
\usepackage{makecell}
\usepackage[capitalise,noabbrev]{cleveref}
\Crefname{figure}{\text{Fig.}}{\text{Figs.}}	
\definecolor{Lightgray}{gray}{0.9}
\usepackage{url}

\newcolumntype{Z}{>{\raggedright\let\newline\\\arraybackslash\hspace{0pt}}X}

\usepackage{bbding}
\usepackage{comment}
\usepackage{caption}
\usepackage{subcaption}
\usepackage{tabularx,ragged2e,siunitx}
\Crefname{figure}{\text{Fig.}}{\text{Figs.}}	
\hypersetup{citecolor=Blue}

\def\tsc#1{\csdef{#1}{\textsc{\lowercase{#1}}\xspace}}
\tsc{WGM}
\tsc{QE}
\tsc{EP}
\tsc{PMS}
\tsc{BEC}
\tsc{DE}
\begin{document}
\let\WriteBookmarks\relax
\def\floatpagepagefraction{1}
\def\textpagefraction{.001}

\shorttitle{Driver Models for Autonomous Vehicles}

\shortauthors{Cheng Wang et~al.}

\title [mode = title]{The Application of Driver Models in the Safety Assessment of Autonomous Vehicles: A Survey}                      



\author[1]{Cheng Wang}
\ead{cheng.wang@ed.ac.uk}
\credit{Conceptualization, Methodology, Visualization, Software, Writing – original draft, Writing – review \& editing.}
\affiliation[1]{organization={School of Informatics, University of Edinburgh},
    %addressline={}, 
    city={Edinburgh},
    % citysep={}, % Uncomment if no comma needed between city and postcode
    postcode={EH8 9AB}, 
    country={United Kingdom}}

\author[2]{Fengwei Guo}
% Email id of the second author
\ead{fengwei.guo@student.tugraz.at}
\credit{Methodology, Visualization, Formal analysis, Writing – original draft, Writing – review \& editing}
\affiliation[2]{organization={Vehicle Safety Institute, Graz University of Technology},
    city={Graz},
    % citysep={}, % Uncomment if no comma needed between city and postcode
    postcode={8010}, 
    %state={},
    country={Austria}}

\author[3]{Ruilin Yu}
% Email id of the third author
\ead{yurl21@mails.jlu.edu.cn}
\credit{Writing – original draft, Writing – review \& editing}

% Forth author
\author[3]{Luyao Wang}
% Email id of the third author
\ead{wangly21@mails.jlu.edu.cn}
\credit{Writing – original draft, Writing – review \& editing}

% Fifth author
\author[3]{Yuxin Zhang}
\cormark[1]
% Email id of the fourth author
\ead{yuxinzhang@jlu.edu.cn}
\credit{Funding acquisition, Project administration, Methodology, Discussion Writing – review \& editing.}

\affiliation[3]{organization={State Key Laboratory of Automotive Simulation and Control, Jilin University},
    city={Changchun},
    % citysep={}, % Uncomment if no comma needed between city and postcode
    postcode={130025}, 
    %state={},
    country={China}}

\cortext[cor1]{Corresponding author}

%\nonumnote{This work was supported in part by the German Federal Ministry for Economic Affairs and Climate Action within the PEGASUS project family “VVM-Verification \& Validation Methods for Automated Vehicles Level 4 and 5”.}

\begin{abstract}
Driver models play a vital role in developing and verifying autonomous vehicles (AVs). Previously, they are mainly applied in traffic flow simulation to model driver behavior. With the development of AVs, driver models attract much attention again due to their potential contributions to AV safety assessment. The simulation-based testing method is an effective measure to accelerate AV testing due to its safe and efficient characteristics. Nonetheless, realistic driver models are prerequisites for valid simulation results. Additionally, an AV is assumed to be at least as safe as a careful and competent driver, which is modeled by driver models as well. Therefore, driver models are essential for AV safety assessment from the current perspective. However, no comparison or discussion of driver models is available regarding their utility to AVs in the last five years despite their necessities in the release of AVs. This motivates us to present a comprehensive survey of driver models in the paper and compare their applicability. Requirements for driver models as applied to AV safety assessment are discussed. A summary of driver models for simulation-based testing and AV benchmarks is provided. Evaluation metrics are defined to compare their strength and weakness. Finally, potential gaps in existing driver models are identified, which provide direction for future work. This study gives related researchers especially regulators an overview and helps them to define appropriate driver models for AVs. 
\end{abstract}

% \includegraphics{figs/grabs.pdf}

%\item Research highlights item 1
%\item Research highlights item 2
%\item Research highlights item 3
% Keywords
% Each keyword is separated by \sep
\begin{keywords}
Autonomous vehicles \sep Driver models \sep Vehicle safety \sep Verification \& validation \sep Safety assessment
\end{keywords}

\maketitle

\section{Introduction} \label{introduction}
Autonomous vehicles (AVs) have been intensively studied in recent years. This significantly drives the evolution of vehicles to a smarter and more intelligent level. As an example of the achievements, Level 2 systems \citep{sae_j3016_taxonomy_2021} have been introduced into the market recently. Although drivers may not use level 2 systems as intended and thus additional risks emerge \citep{kim_is_2022}, the level 2 systems could in principle increase driving safety and comfort by controlling both the longitudinal and lateral motion of a vehicle. To further increase autonomy, Level 3 systems are supposed to be the goal for the next stage by shifting the entire dynamic driving tasks (DDT) to the systems themself in a predefined operational design domain (ODD) \citep{sae_j3016_taxonomy_2021}. Before bringing Level 3 systems into the market, corresponding verification and validation (V\&V) procedures are essential to prove their safety. However, numerous known and unknown unsafe scenarios exist due to the complexity of the open world and the system itself, as well as the increasing interaction with road users, the safety validation without an explicit stop criterion seems infeasible. Thus, the typical question "\textit{How safe is safe enough}" \citep{liu_how_2019} arises. 

To answer this question, defining safety goals for autonomous vehicles (AVs) becomes imperative. The safety goals differ across AV functions and systems. A safety goal in ISO 26262 \citep{iso_iso_2018} is defined as a low and acceptable residual risk for electric/electronic systems. In contrast, the accident rate per kilometer is described in ISO 21448 \citep{iso_iso_2022} as the validation target for AVs. Recently, the newly released EU regulation 2022/1426 \citep{EU_regulation_2022} defines the accident rate per hour as the validation target to avoid the influence of speed. The accident rate of human drivers given a specific confidence interval is usually determined by statistic analysis \citep{oboril_mtbf_2022}. In this way, the derived validation target is the goal that an AV is expected to achieve. A similar concept is proposed in \citep{kauffmann_positive_2022,iso_isotr_4804_road_2020} where a positive risk balance (PRB) compared with human driving performance is suggested prior to the launch of AVs. 

However, it is difficult to prove that any given system in each specific ODD has achieved the validation target. This is not only because finding the baseline accident rate for each specific ODD is a cumbersome task, but also because the amount of testing to validate the accident rates for any given system can be insurmountable \citep{kalra2016driving}. Further, large-scale testing in the actual ODD, before another type of validation, may put the rest of the traffic at unreasonable risk \citep{wang2021online}. Those are some of the issues raised, that make the use of driver models an important tool, for setting the requirements, and validating AV technologies. Motivated by this, the UNECE released regulation No.157 \citep{un_ece_regulation_2021} for L3 Automated Lane Keeping Systems (ALKS) with a maximum driving speed of 130 km/h. This regulation suggests that an AV's performance shall be ensured at least to the level at which a \textbf{careful and competent human driver} could minimize the risks. To depict a competent and careful human driver, two driver performance models are introduced. They are a reaction time-based driver model (the Japanese driver model \citep{experts_of_japan_competent_2020}) and a fuzzy safety model (FSM \citep{mattas_driver_2022}), respectively. Cut-in, cut-out, and deceleration scenarios with various parameter combinations are simulated to determine under which situation a collision occurs and which does not. The simulation results are used as a guide when testing AVs in the same test scenarios to identify whether AVs can avert some collisions or cause more collisions.  

\begin{figure*}[!ht]
  \begin{minipage}[t]{.3\linewidth}
    \includegraphics[width=\linewidth]{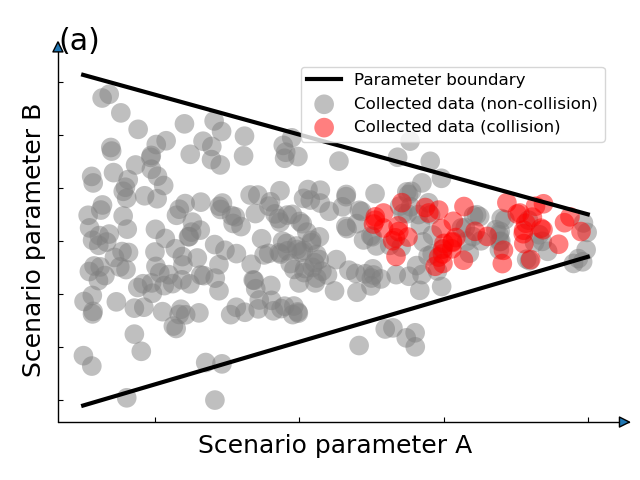}%
  \end{minipage}\hfil
  \begin{minipage}[t]{.3\linewidth}
    \includegraphics[width=\linewidth]{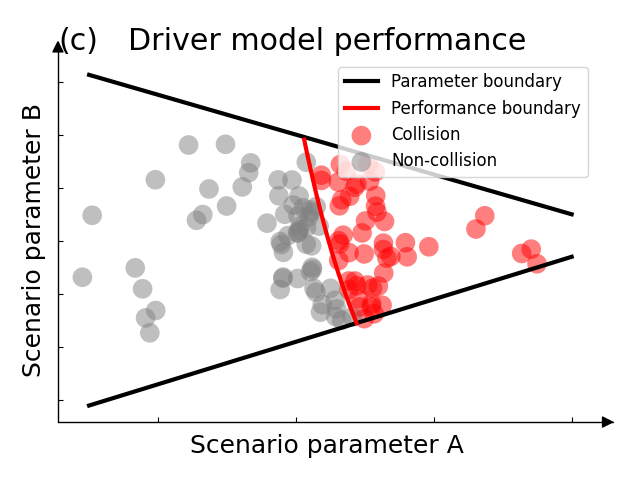}%
  \end{minipage}\hfil
  \begin{minipage}[t]{.3\linewidth}
    \includegraphics[width=\linewidth]{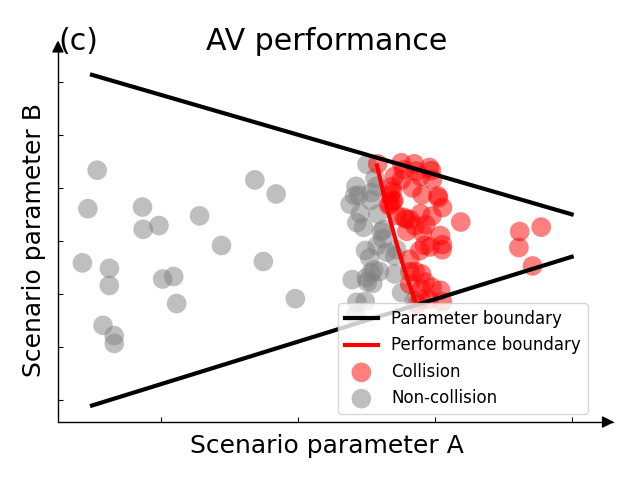}%
  \end{minipage}%
  \caption {\label{fig:driver_model} The application of driver models for AV safety verification. (a) represents the determination of reasonably foreseeable parameter ranges using collected data; (b) illustrates the collision avoidance capability of a driver model using adaptive sampling within the parameter ranges; (c) depicts an AV's collision avoidance capability using adaptive sampling within the parameter ranges. The AV is supposed to perform better than the reference driver model, as the comparison between (b) and (c) shows. } 
\end{figure*}

\cref{fig:driver_model} illustrates the concrete application process of a driver model to determine its collision avoidance capability. First, reasonably foreseeable parameter ranges \citep{nakamura_defining_2022} are determined by the collected data from the real world. They are defined as likely occurring scenarios within a specific ODD and period. According to risk acceptance and relevant exposure, the limits of reasonably foreseeable parameter ranges vary. Subsequently, sampling techniques such as uniform sampling and adaptive sampling \citep{wang_safety_2022} are applied to generate concrete scenarios \citep{menzel_scenarios_2018} based on the parameter ranges and correlations. Finally, the collision avoidance capability of a driver model is determined by distinguishing collision and non-collision scenarios executed in simulations.

In addition, driver models are essential for the simulation-based method \citep{weber_simulation-based_2020,espineira_realistic_2021} for AV V\&V. In the simulation-based method, AVs are tested in a virtual environment with surrounding vehicles powered by driver models. Due to the huge test effort involved in the distance-based method, shifting V\&V to simulations seems an inevitable choice to escape from the “approval trap”. As indicated in \citep{wachenfeld_release_2016}, 6.62 billion test kilometers must be driven to prove that an AV is approximately twice as good as human-driven vehicles at a significant level of 5\% according to the fatal accidents from the German Federal Statistical Office. If non-careful and competent drivers are excluded from the statistics, the test kilometers would further increase. With this fact in mind, projects like VV-Method \citep{vvm_verification_2022} and SET Level \citep{set_level_set_2022} are initiated to develop a seamless chain of reasoning for the proof of safety and a holistic tool chain. Both projects emphasize the importance of the simulation-based method. 

Even in the simulation-based method, four subcategories \citep{pears_consortium_prospective_2021} are recommended by P.E.A.R.S (an open consortium to harmonize the prospective effectiveness assessment of active safety systems by simulation):
\begin{enumerate}
  \item Direct usage of real-world cases (i.e., reconstructed crash or field data) without any changes.
  \item Usage of real-world cases plus varying the initial values by means of distribution.
  \item Deriving scenario mechanisms and distributions from real-world cases and selecting a low number of representative cases.
  \item Deriving scenario mechanisms and distributions from real-world cases and applying sampling to generate multiple cases.
\end{enumerate}
The last subcategory is suggested by Fries et al.\citep{fries_driver_2022} and Kaufmann et al. \citep{kauffmann_positive_2022} because this approach does not directly rely on real-world cases but establishes the link to them via distributions. Since no predefined trajectories are available, a driver model is required to update the state of surrounding vehicles with given initial states during the simulation. According to the level of detail to represent a traffic flow, microscopic, mesoscopic, and macroscopic simulation models are distinguished \citep{ferrara_microscopic_2018,hoogendoorn_state---art_2001}. In microscopic simulation, the space-time behavior of vehicles and drivers as well as their interactions are modeled at a high level of detail. Low dynamics are described in aggregate terms using probability distribution functions and the dynamics of these distributions are governed by individual drivers’ behavior in mesoscopic simulation. In macroscopic simulation, traffic flow is represented as large road networks, measured in terms of characteristics such as speed, flow, and density.

\begin{comment}
\begin{enumerate}
  \item {\bf microscopic} simulation models: the space-time behavior of vehicles and drivers as well as their interactions are modeled at a high level of detail;
  \item {\bf mesoscopic} simulation models: the traffic flow dynamics are described in aggregate terms using probability distribution functions and the dynamics of these distributions are governed by individual drivers’ behavior;
  \item {\bf macroscopic} simulation models: traffic flow is represented as large road networks, measured in terms of characteristics such as speed, flow, and density.
\end{enumerate}
\end{comment}

Apparently, driver models for AV safety assessment can be applied to microscopic and macroscopic simulations. Specifically, driver models are helpful to model the driving behavior of each individual vehicle to evaluate AV's performance to handle driving tasks in short-term concrete scenarios. Similarly, they can be applied in macroscopic simulation for analyzing the impact of AVs on traffic flow. 

As a result, driver models play a vital role in the safety assessment of AVs. Nevertheless, since AV certification is challenging and achieving PRB with driver models as references remain under exploration, little review regarding their application for AV verification is currently available despite the fact that the driver model has been studied for decades. In the recent five years, Singh and Kathuria \citep{singh_analyzing_2021} discussed the study of driving behavior using naturalistic driving data. Similarly, a review was conducted in \citep{gouribhatla_vehicles_2022} to analyze the effect of advanced features such as adaptive cruise control (ACC) on driver behavior. Park and Zahabi \citep{park_review_2022} presented a review of human performance models focusing on human cognition and interactions with in-vehicle technology. A survey on car-following models was performed in \citep{matcha_simulation_2020,ahmed_review_2021} without covering AV safety assessment, which is the same as the reviews made five years ago, e.g., Rahman et al. \citep{rahman_review_2013}  and Moridpour et al. \citep{moridpour_lane_2010} gave a review of lane-changing models in 2013 and 2012, respectively.

Therefore, we are motivated to present a detailed and holistic review of currently available driver models in terms of their application for AV safety assessment by considering the following three research questions:
\begin{itemize}
  \item \textbf{RQ1}: \textit{what are the requirements on the driver models for AV safety assessment?}
  \item \textbf{RQ2}: \textit{what driver models are available in this context considering the requirements?}
  \item \textbf{RQ3}: \textit{What kind of driver models are appropriate for what kind of AV safety assessment tasks?}
\end{itemize}
To the best of the authors' knowledge, no survey on driver models focuses on their application in AV safety assessment. Additionally, lateral avoidance is also an important maneuver, apart from the models for longitudinal braking when confronting a critical situation, appropriate driver models for lateral avoidance are also included, whereas few surveys about driver models take this into account. After answering these three research questions, we discuss the current gaps and future research directions. Therefore, our contributions to the work are as follows:
\begin{itemize}
  \item requirements on driver models for AV safety assessment are derived, which facilitates AV developers to develop their driver models for AV testing;
  \item a comprehensive survey of driver models is presented. This gives developers and regulators an overview of the current status;
  \item appropriate driver models for AV safety assessment are compared based on our proposed metrics. Thus, a selection of appropriate models for related researchers is possible by our comparison;
  \item a thoughtful discussion of possible driver models for AV certification is given to indicate future working directions.
\end{itemize}

Section \ref{Functions and requirements} primarily addresses the requirements for driver models in terms of their application in AV safety assessment, as well as the scope of relevant driver models. Based on the determined scope, the driver models aiming at modeling driver behavior in simulations are elaborated in Section \ref{Driver models for simulations}. Section \ref{Driver models as references} deals with the collision avoidance capability of driver models for benchmark applications. Subsequently, the applicability of driver models for AV safety assessment is highlighted in Section \ref{Applicability}. Based on the analysis, a discussion is conducted and limitations are identified in Section \ref{Discussion}. Lastly, the conclusion and future works are summarized in Section \ref{Conclusion and future work}.

\section{Functions and requirements} \label{Functions and requirements}
In this section, we first explore where driver models are helpful for AV V\&V. Based on this, we analyze what requirements those specific needs impose on the driver models. These derived requirements serve as a foundation for the detailed description of currently available driver models in the following sections and indicate potentially the current gaps for future research. 

\subsection{Functions}
\begin{figure}[!ht]
\centering
  \begin{minipage}{.75\linewidth}
    \centering
    \includegraphics[width=\linewidth]{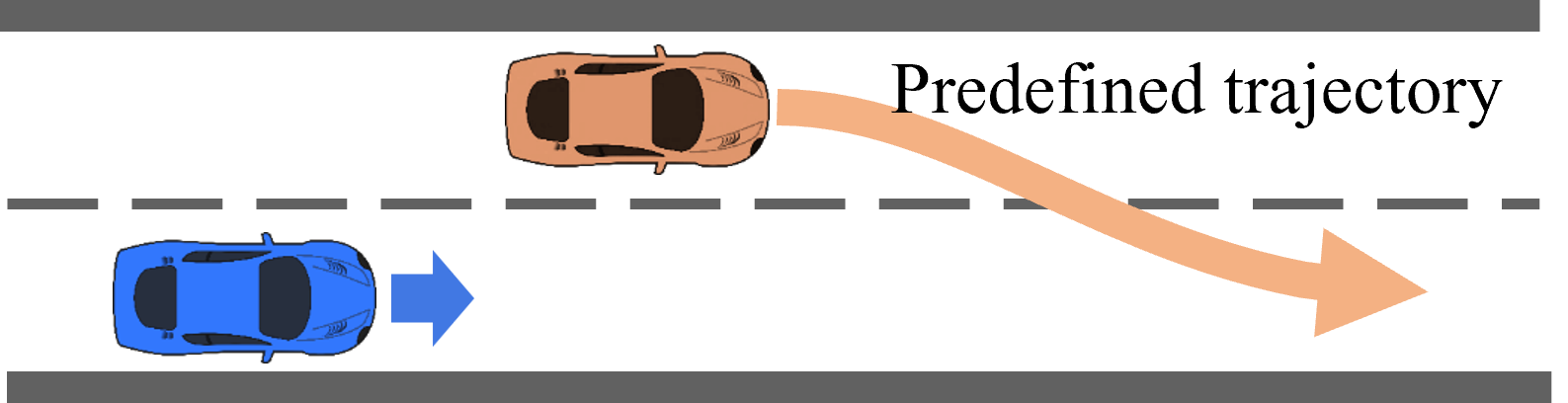}%
    \subcaption{}\label{subfig-1}
  \end{minipage}\hfil
  \begin{minipage}{.75\linewidth}
   \centering
    \includegraphics[width=\linewidth]{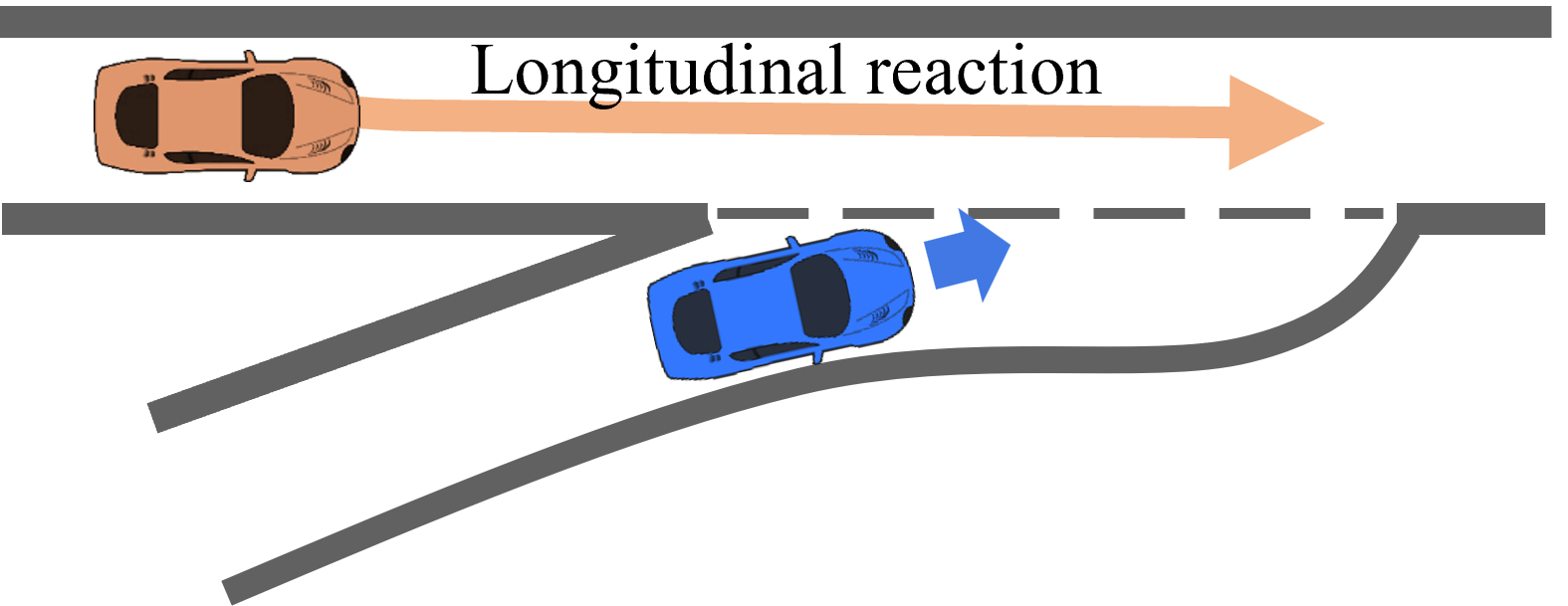}%
    \subcaption{}\label{subfig-2}
  \end{minipage}\hfil
  \begin{minipage}{.75\linewidth}
    \centering
    \includegraphics[width=\linewidth]{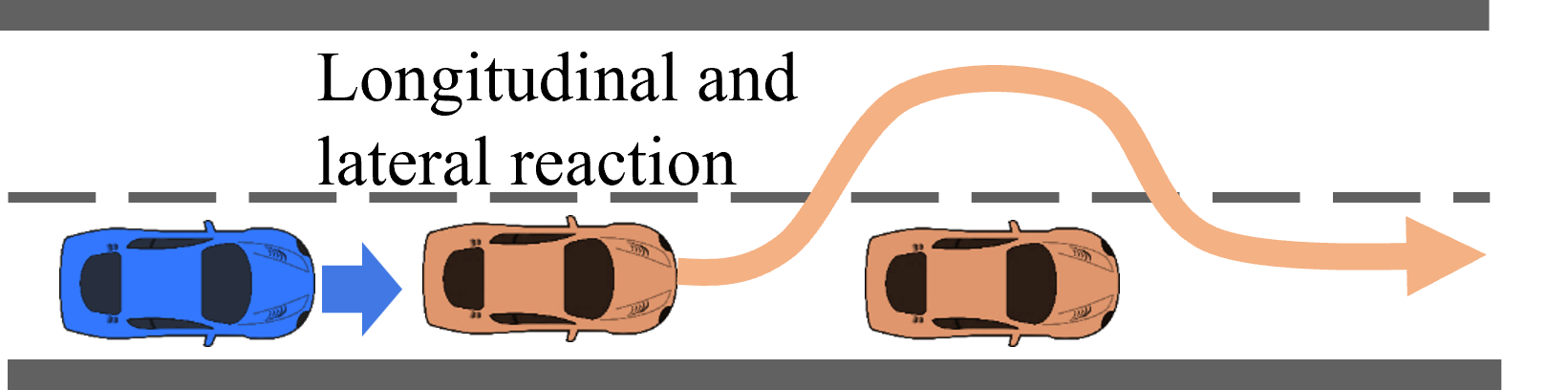}%
    \subcaption{}\label{subfig-3}
  \end{minipage}\hfil
    \begin{minipage}{.75\linewidth}
    \centering
    \includegraphics[width=\linewidth]{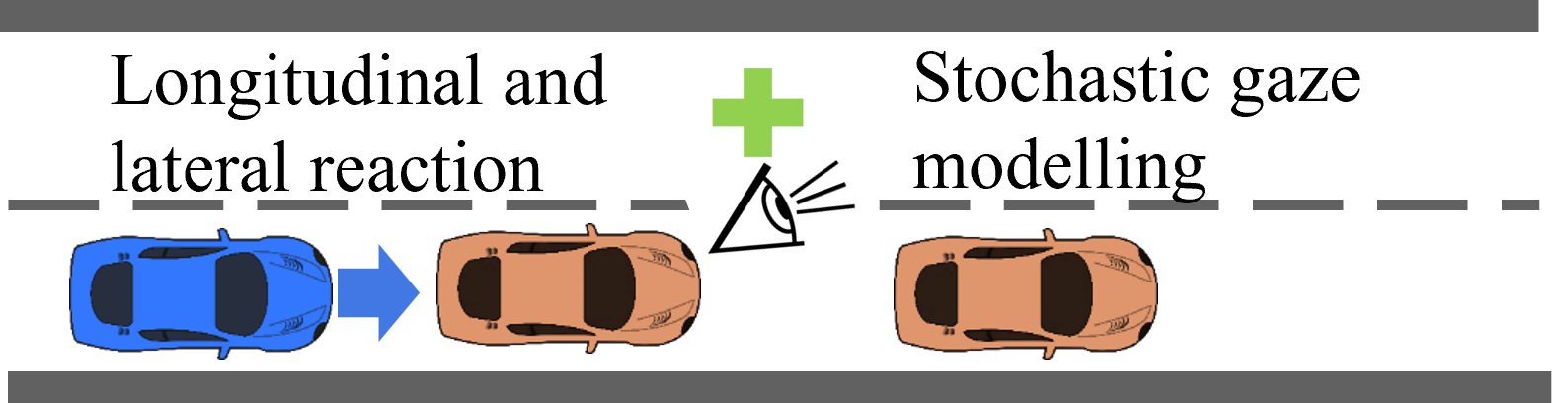}%
    \subcaption{}\label{subfig-4}
  \end{minipage}%
  \caption {\label{fig:simulation} The application of driver models in simulation-based testing. (a) trajectories of surrounding vehicles are predefined; (b) the behavior of surrounding vehicles is longitudinally controlled; (c) both lateral and longitudinal behavior of surrounding vehicles are controlled; (d) the information acquisition process is included in driver models besides longitudinal and lateral control. } 
\end{figure}

Regarding the functions of driver models in AV V\&V, simulation-based testing, and reference models are the two places where driver models are typically applied.
\cref{fig:simulation} presents four different functions of driver models in simulation-based testing. 
Note, we assume a low penetration rate of AVs. The surrounding vehicles are driven by humans. Therefore, the driver models discussed in the paper focus on modeling human drivers' behavior rather than AVs'.
In \cref{subfig-1}, the trajectories of surrounding vehicles are predefined, which are typically retrieved from naturalistic driving data, traffic accident data, etc., or are generated by, for example, a constant velocity model given their initial states. This kind of driver model is simple but suffers from unrealistic driving behavior.

Many car-following models are proposed, such as the Intelligent Driver Model (IDM) \citep{treiber_congested_2000}, with the goal to enable longitudinal interaction between vehicles. As illustrated in \cref{subfig-2}, the surrounding vehicle shows its "politeness" to enable the merging of the ego vehicle. By calibrating model parameters using naturalistic driving data, the driver models are statistically shown to be capable of recreating realistic traffic flow \citep{sharma2019more}. However, car-following models do not fully portray driving behavior in the real world. Necessary lateral driving behavior is also common in daily driving situations. Therefore, lane-changing models are studied, which further increase the interaction level. \cref{subfig-3} shows an example, where the surrounding vehicle overtakes a slow leading vehicle. This makes the decision-making of the ego vehicle face a more realistic driving challenge, and thus more valuable to test AVs. 

Recently, the stochasticity of information processing and situation understanding are considered when modeling driver models in order to simulate inattentive or distracted driving behaviors of human drivers \citep{kitajima_nationwide_2022}.
This driver model is categorized as {\bf cognitive models} \citep{tattegrain-veste_computational_1996}, in which the internal processes and states that produce the behavior are modeled. {\bf Predictive models} \citep{tattegrain-veste_computational_1996}, on the other hand, attempt to simulate the driver behavior without necessarily considering the underlying processes that lead to the behavior. The cause-and-effect relationships between the behavior and the external factors are ignored, which results in limited predictive capabilities \citep{siebke_what_2022}. Thus, cognitive models aim to model the entire reaction process of a human driver when dealing with driving situations.

In addition, driver models can also be utilized as a reference for AVs. For instance, an AV is to be blamed if it causes an accident, while a careful and competent driver does not in the same driving scenario \citep{koopman_reasonable_2023}. To model a reference driver model for AVs, the safety performance of the driver model itself shall be convincing and acceptable. Otherwise, the safety performance of AVs may still be unsatisfying if a less competent driver model is taken as the reference. Since drivers' peak performance capabilities are usually elicited in critical scenarios, while routine scenarios elicit typical (not necessarily the best) behavior \citep{shinar_review_2011}, a reference driver model is developed based on driver performance data in critical scenarios. An example is the Japanese driver model \citep{experts_of_japan_competent_2020}, where trained drivers are employed to conduct emergency braking experiments to determine model parameters. By analyzing the collision avoidance capability of the reference driver model in the test scenarios derived from an AV's ODD, we can determine whether the AV would avoid more collisions or even cause more collisions in the same test scenarios. While the test scenarios can be simulated by using driver models for simulations, a joint application of these two types of models is possible. 

Consequently, driver models play a vital role in the safety assessment of AVs. In simulation-based testing, various driving scenarios can be simulated with the help of predictive driver models. In particular, driving behavior with "surprises" can be created by cognitive models, which could result in some known or even unknown critical scenarios for AVs. Thus, predictive and cognitive driver models are valuable to test AVs. In addition, driver models can also be regarded as references when assessing the safety performance of AVs, if the driver models could represent careful and competent human drivers. By comparing the safety performance of a reference driver model and an AV in the same test scenario, the safety evaluation of the AV is possible.

%Depending on the scale of the simulation, predictive models are suitable for replicating human-like trajectories in concrete scenarios, whereas cognitive models are more appropriate for large-scale simulations in order to generate human-like driving errors randomly. Conversely, the driver model as a baseline for AV safety verification shall reflect the peak human driving capability in order to obtain the performance boundary.

\subsection{Requirements} \label{requirements}
In order to provide sound and credible test results, a high-validity simulated environment is usually required. To this end, the driver behavior in simulation shall be human-like to simulate realistic driving scenarios in the real world, and the interaction between various road users shall be reasonable. Therefore, \textbf{interaction-aware} and \textbf{human-like} driver models are demanded in simulation-based testing. We emphasize that human-like trajectories do not necessarily mean interaction-aware. For instance, a well-calibrated driver model could possibly deliver human-like trajectories but may fail to interact with surrounding vehicles properly.
Note, when it comes to interaction, we refer to driver models' proactive reaction to other vehicles. How the AV will react to the surrounding vehicles is out of scope.
Additionally, different \textbf{driver characteristics} are essential to be modeled because aggressive and defensive drivers behave differently and thus result in scenarios with different criticality. 
Finally, \textbf{cognitive capabilities} are essential for the driver models to consider the situation understanding process, with the aim to simulate possible human errors while during.

\begin{figure}[t] 
\includegraphics[width=0.45\textwidth]{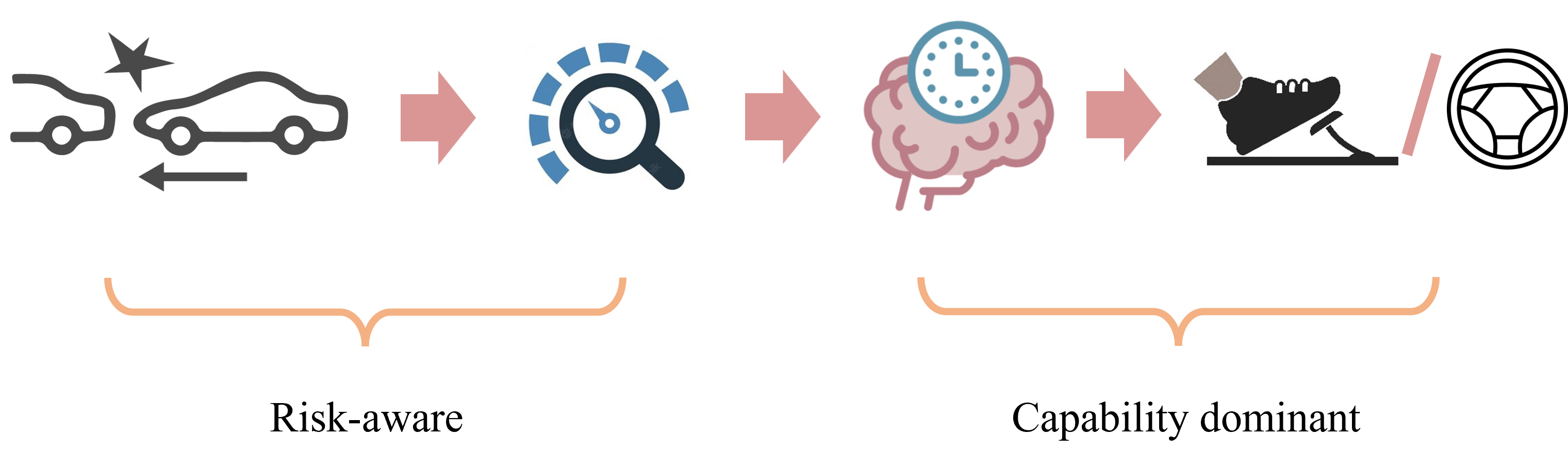}
\centering
\caption {\label{fig: reference} Risk-aware and capability dominant driver models as references for AV safety assessment. } 
\end{figure}

How to consider "careful" and "competent" is vital for the reference driver models. Careful can be interpreted as timely risk awareness. A careless driver may lead to a collision due to late risk recognition, which will make an AV risky based on this reference. In contrast, a too-careful driver could be too conservative, which undermines the mobility of an AV. Therefore, timely \textbf{risk-aware} decision-making is fundamental for reference driver models. Competent, on the other hand, focuses on excellent driving capability. While driving capability is not fully demonstrated in nominal scenarios, driving performance in critical scenarios is a more appropriate measure. Nevertheless, the driving capability among drivers varies, for instance, the driving performance of a less skilled driver in critical scenarios could still be an unsatisfying reference for AVs. Hence, the experiment data from skilled and experienced drivers to build a reference driver model is more compelling. As a result, \textbf{capability dominant} driver models are needed to articulate competence. \cref{fig: reference} shows the two fundamental requirements of a careful and competent driver when dealing with critical scenarios.

\subsection{Classification}
\begin {figure*}[!hbtp]
\centering
\begin{tikzpicture}[
squared_node/.style={rectangle, draw=black!60, fill=gray!5, rounded corners=3, very thick, minimum width = 2.5cm, minimum height = 1cm},
secondary_node/.style={rectangle, draw=black!60, fill=orange!5, rounded corners=3, very thick, minimum size=5mm},
third_level_node/.style={rectangle, draw=black!60, fill=red!5, rounded corners=3, very thick, minimum size=5mm},
arrow/.style = {thick,->,>=stealth}
]
%Nodes
\node[squared_node]      (root)         {Driver models};
\node[squared_node]      (simulation)       [below of=root, xshift=-4cm, yshift=-0.6cm] {For simulation};
\node[squared_node]      (reference)       [below of=root,xshift=4cm, yshift=-0.6cm] {As references};

\node[secondary_node]    (car-following)       [below of=simulation, xshift=-2.8cm,  yshift=-0.6cm, text width=2.0cm, align=center] {Car-following};
\node[secondary_node]    (lane-changing)       [right of=car-following, xshift=1.6cm, text width=2.4cm, align=center] {Lane-changing};
\node[secondary_node]    (cognitive)       [right of=lane-changing, xshift=1.6cm, text width=2.0cm, align=center] {Cognitive};
\node[secondary_node]    (avoidance by steering)       [below of=reference,  yshift=-0.6cm, text width=2.1cm, align=center] {Avoidance by steering};
\node[secondary_node]    (avoidance by braking) [left of=avoidance by steering, xshift=-1.8cm, text width=2.5cm, align=center] {Avoidance by braking};
\node[secondary_node]    (avoidance by steering and braking)       [right of=avoidance by steering, xshift=1.85cm, text width=2.7cm, align=center] {Avoidance by steering \& braking};

\node[third_level_node]    (car-following-Rule-based)       [below of=car-following, yshift= -0.6cm, text width=2cm, align=center] {\citep{gipps_behavioural_1981,treiber_congested_2000,jia_development_2001,xu_development_2007}
};

\node[third_level_node]    (lc-Rule-based)       [below of=lane-changing, yshift= -0.8cm, text width=2.4cm, align=center] {
\citep{gipps_model_1986,toledo_modeling_2003,schakel_integrated_2012,xie_data-driven_2019,wang_reinforcement_2018} };

\node[third_level_node]    (cognitive-based)       [below of=cognitive, yshift= -0.04cm, text width=2.0cm, align=center] {
\citep{siebke_report_2021,fries_driver_2022} 
};

\node[third_level_node]    (b-braking-based)       [below of=avoidance by braking, text width=2.6cm, yshift=-0.8cm, align=center] {
\citep{markkula_farewell_2016,un_ece_regulation_2021,shalev-shwartz_formal_2017,mattas_driver_2022}
};

\node[third_level_node]    (b-steering-based)       [below of=avoidance by steering, text width=2.1cm, yshift=-0.8cm, align=center] {
\citep{feher2020hierarchical,park_emergency_2021,wurts2021collision,dona_towards_2023}
};

\node[third_level_node]    (braking-steering-based)       [below of=avoidance by steering and braking, text width=2.7cm, yshift=-0.62cm, align=center] {
\citep{jurecki_driver_2009,li_emergency_2022,engstrom_modeling_2022}
};

%Lines
\draw [arrow] (root) -- (simulation);
\draw [arrow] (root) -- (reference);
\draw [arrow] (simulation) -- (car-following);
\draw [arrow] (simulation) -- (lane-changing);
\draw [arrow] (simulation) -- (cognitive);
\draw [arrow] (reference) -- (avoidance by braking);
\draw [arrow] (reference) -- (avoidance by steering);
\draw [arrow] (reference) -- (avoidance by steering and braking);
\end{tikzpicture}
\caption {\label{fig:Architecture} The review scope of the paper and the classification of driver models. Car-following, lane-changing, and cognitive models are discussed in terms of their applications in testing AVs in simulations. For driver models as references, braking, steering, and a combination of both for collision avoidance are elaborated.} 
\end{figure*}
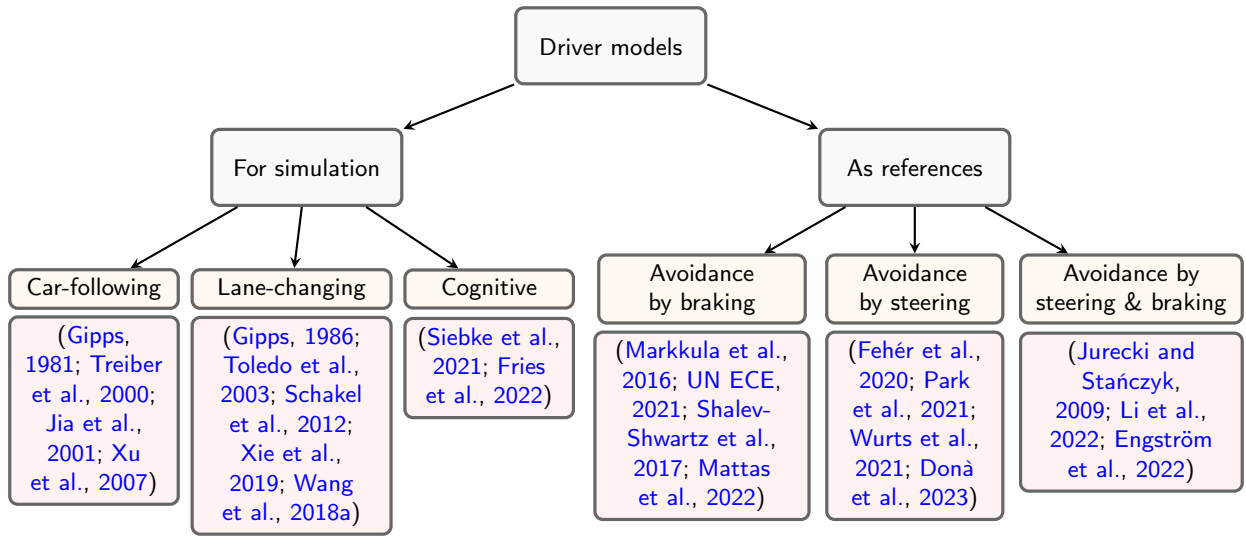

The derived requirements provide a solid basis for exploring existing driver models and identifying their gaps. The identified driver models by the authors to cover the aforementioned requirements are categorized in \cref{fig:Architecture}.

With respect to  {\bf driver models for simulations}, we categorize the identified driver models for their suitability to meet the requirements of interaction-aware, human-like, driver characteristics and cognitive capabilities. This includes existing car-following, lane-changing, and cognitive models. The human cognition process is included in cognitive models to model driving errors, whereas car-following and lane-changing models belong to predictive models by focusing on the driver behavior itself. For each type of model, we further classify the models according to their model characteristics.

In addition to driver models for simulation, {\bf driver models as references} are elaborated in terms of their risk-aware considerations and driving capability modeling processes. These models attempt to model a driver's reaction to an imminent situation. Representative examples from this category are the Japanese model \citep{experts_of_japan_competent_2020} defined in UNECE Regulation No.157 \citep{un_ece_regulation_2021}. The reference driver models in the regulation consider braking as the strategy to avoid collisions. As an extension, reference driver models that include evasive steering maneuvers are essential, as the evasive steering maneuver is also a decent action for collision avoidance if free space is available on the side \citep{eckert_emergency_2011}.
Furthermore, simultaneous steering and braking maneuvers can reduce the braking distance further if a collision is inevitable \citep{choi_simultaneous_2017}. Therefore, we review not only braking models, but also steering models and braking \& steering models in the paper. Consequently, the review benefits developers and regulators to develop a comprehensive driver model for AV safety verification. For each possible maneuver, we further divide them into different categories considering their modeling processes.

%...

\section{Driver models for simulations} \label{Driver models for simulations}
In this section, we present existing driver models for simulation-based testing and describe which requirements discussed in \cref{requirements} are considered in the models. In particular, car-following and lane-changing models focusing on different types of interaction are distinguished. 

%driver models consist of stimulus, reaction, and execution. 

\subsection{Car-following models} \label{car_following_models}
Car-following models are common driver models for modeling longitudinal interaction. Various car-following models, such as Gipps model \citep{gipps_behavioural_1981}, Newell model \citep{newell_simplified_2002}, and Optimal Velocity (OV) model \citep{bando_dynamical_1995} have been developed since the Gazis-Herman-Rothery (GHR) model \citep{chandler_traffic_1958} was proposed. To simulate the interaction, stimulus and reaction are considered in car-following models. The relative state between the preceding and following vehicles is usually used as a stimulus, while the deceleration of the following vehicle is the reaction. For instance, the GHR model \citep{chandler_traffic_1958} utilizes relative speed as a stimulus item, while the Intelligent Driver Model (IDM) model \citep{treiber_congested_2000} does not define an explicit stimulus item, but uses the state of the preceding vehicle directly.

Unlike the IDM model, psychological-physical car-following models aim to define a psychologically safe distance as a stimulus account. For instance, Wiedemann introduced the term "perceptual threshold" to define the minimum value of a stimulus that the driver can perceive and respond to \citep{wiedemann_simulation_1974}. Once the following driver believes that the relative distance to the preceding vehicle is less than the psychological safety distance, the driver starts to slow down. Conversely, the driver accelerates to reach the psychological safety distance. Considering the way the brain estimates the collision time, Andersen et al. \citep{andersen_optical_2007} proposed the Driving-by-Visual-Angle (DVA) model, which uses the visual angle and its change rate as variables for the driver to make acceleration/deceleration decisions.
 
However, it is difficult for psychological-physical models to find a balance between simplicity and performance due to the complex perceptual processes of the drivers. Cellular Automaton (CA) is a promising approach to address this challenge. It is defined as a dynamical system that evolves in discrete time dimensions according to certain local rules in a cellular space composed of cells with discrete and finite states. The empty cells in front and the current velocity of the following vehicle are coded as stimuli. Since the model developed by Nagel and Schreckenberg (NaSch) \citep{nagel_cellular_1992}, many improved CA-related driver models are proposed such as considering driver characteristics \citep{zamith_new_2015, malecki_multi-agent_2023}.

%Despite being widely used in traffic flow simulation, the CA model falls short of meeting the accuracy required by vehicle simulation. This is due to the contradiction between simplicity and authenticity inherent in the CA algorithm. Recently, driver models incorporating driver characteristics are studied. Liao \citep{liao_car-following_2019} proposed different theoretical models for three traffic states, considering drivers' varying driving styles in different traffic conditions. Afterward, numerical tests were conducted to verify the safety, stability, comfort, fuel economy, and consistency of these models with various driving styles in various scenarios. Similarly, Chen \citep{chen_investigating_2020} investigated intrinsic long-term driving characteristics and their short-term changes using drivers experiencing external stimuli, and proposed a long and short-term driving (LSTD) model incorporating these changes into the car-following model. The model was evaluated using the NGSIM dataset \citep{us_department_of_transportation_next_2016}. 

With the advent of big data and the rapid improvement of data collection technology, high-precision and large-sample trajectory data can be obtained easily, stimulating the development of data-driven car-following models. Instead of adhering to various theoretical assumptions and pursuing mathematical derivations in a strict sense, data-driven models use non-parametric methods to mine the intrinsic information of trajectory data and build car-following models with high prediction accuracy.
For instance, backpropagation (BP) neural networks\citep{jia_development_2001}, radial basis function neural network \citep{xu_development_2007,zhou_application_2009}, and fuzzy neural networks \citep{huang_use_1999,ma_neural-fuzzy_2006,li_research_2007} were proposed to model car-following behavior. However, the generalization of these models in unseen situations is usually limited.

 Support vector regression is a regression algorithm based on the support vector machine framework. It can be used for regression fitting of trajectory data. This method follows the principle of structural risk minimization and theoretically has stronger data learning and generalization abilities than artificial neural networks. An exemplary application is the model studied by Zhang et al. \citep{zhang_study_2018}. Based on the assumption that drivers tend to exhibit similar driving behaviors when facing the same driving scenario, He et al. \citep{he_simple_2015} searched the \textit{K} similar historical driving scenarios for the most likely driving behaviors, which were then used as model output to generate a KNN (\textit{K}-nearest-neighbor) car-following model. Compared to other data-driven models with opaque structures, the KNN model has a clearer modeling structure and is more understandable.

 Deep learning (DL) models, compared to traditional neural network models, usually have multiple hidden layers and a correspondingly huge number of neuronal connection weights, thresholds, and other parameters. Various DL-based car-following models have been concentrated in the past five years \citep{zhou_recurrent_2017,wang_capturing_2018,lee_integrated_2019,liu_learning-based_2022}. For instance, both Zhou et al. \citep{zhou_recurrent_2017} and Wang et al. \citep{wang_capturing_2018} proposed car-following models based on recurrent neural networks (RNN) by taking continuous historical time series and vehicle dynamic data as input, while the output is the desired speed for the following vehicle. The results show that their models perform well in predicting the trajectory of the following vehicle. 

However, the high accuracy of DL models comes at the expense of data dependency, high computational costs, and poor generalization. Deep Reinforcement Learning (DRL) addresses these issues to some extent. Zhu et al.\citep{zhu_human-like_2018} used the difference between simulated speed and observed speed as the reward function and considered a 1 s reaction delay to build a car-following model. The model reproduced human-like car-following behavior and showed better generalization ability, as the agent learned decision-making mechanisms from the training data, rather than parameter estimation through data fitting. As an extension, Hart et al. \citep{hart_formulation_2021} incorporated the idea of driving styles in the reward function to simulate different driver characteristics.

Both the traditional analytical and recent data-driven models can be applied in simulations to generate the car-following behavior of surrounding vehicles to test AVs. The analytical models are simple and interpretable, while the data-driven models show superiority in modeling human-like driving behavior and driver characteristics. Depending on the training data, data-driven models could also incorporate careless or distracted driving behavior to consider cognitive processes. Generally, data-driven models show a promising trend.

\subsection{Lane-changing models}
Another sort of driver model required in simulations to provide more diversified traffic scenarios for testing AVs is lane-changing models. From the interaction perspective, free, cooperative, and forced lane-changing are proposed \citep{hidas_modelling_2002,hidas_modelling_2005}. In cooperative and forced lane-changing, the follower slows down reluctantly or willingly to create enough space for the lane changer to insert.

The stimulus, like in car-following models, is the first stage in determining lane-changing maneuvers. However, the stimulus is complicated in lane-changing models, since mandatory lane-changing (MLC) and discretionary lane-changing (DLC) \cite{yang_microscopic_1996,toledo_modeling_2003} exist. MLC happens when the driver must leave the current lane (e.g., to use an off-ramp or avoid a lane blockage), and DLC happens when the driver performs a lane change to improve driving conditions (e.g., to increase the desired speed in the case of a slow leading vehicle).

The Gipps model \citep{gipps_model_1986} is a type of rule-based lane-changing model, which considers the necessity, desirability, and safety when deciding lane-changing. Factors that affect lane-changing are predefined and their importance is evaluated deterministically. Three zones depending on the distance to the intended turn are defined to govern the driver's behavior for the intended lane-changing. More specifically, a desired speed is kept if the intended turn is far away, while lane changes to the turning lanes or adjacent lanes are considered in the middle zone. When the intended turn is close, the driver focuses on keeping the correct lane and ignores gaining other advantages. Due to the clearly structured triggering conditions, the model has been applied in several traffic simulations \citep{christen_driver_2008,casas_traffic_2010}. However, the variability in individual driver behavior \citep{rahman_review_2013}, parameter estimation \citep{toledo_modeling_2003}, and applicability in congested scenarios \citep{moridpour_lane_2010} are not addressed.

Yang and Koutsopoulos \citep{yang_microscopic_1996} defined four steps to model a lane-changing maneuver: the decision to consider a lane-changing, the choice of the target lane, the search for an acceptable gap, and the execution of the change. Different from the Gipps model, the initiation of an MLC is described with a probability that depends on the distance to the intended turn. Although driver characteristics are modeled to some degree, parameter estimation and validation of the model are missing. Afterward, Ahmed's model \citep{ahmed_models_1996,ahmed_modeling_1999} also considers lane-changing probabilistically. The probability of MLC and DLC is calculated in a discrete choice framework. However, a rigid separation between MLC and DLC could be unrealistic in some scenarios because once the MLC is activated, other considerations such as DLC are ignored.  

Therefore, Toledo \citep{toledo_modeling_2003} developed an integrated probabilistic lane-changing model in which MLC and DLC can take effect simultaneously. To evaluate the model, a comparison between separate and integrated MLC \& DLC was performed. The results demonstrated the importance of incorporating trade-offs between MLC and DLC into a lane-changing model.

The minimizing overall braking induced by lane change (MOBIL) \citep{kesting_general_2007} model, on the other hand, measures both the attractiveness of a given lane (i.e., its utility) and the risk associated with lane changes. The reaction is a single-lane acceleration. When a lane change is considered, it is assumed that a driver makes a trade-off between the expected advantage and the disadvantage imposed on other drivers. The advantages are measured by the difference in the accelerations after and before the lane change, while the disadvantages are quantified by the deceleration imposed on the lag vehicle. The MOBIL model has the advantage of transferring the assessment of the traffic situation to the acceleration function of the car-following model, allowing for a compact and general model formulation with only a few additional parameters. Nevertheless, empirical justification, model calibration, and validation remain unaddressed. 

The lane-changing model with relaxation and synchronization (LMRS) \citep{schakel_integrated_2012} is another example to integrate three different incentives including route following, speed gaining, and right keeping into a single desire. By comparing the single desire with three predefined thresholds, no lane-changing, free lane-changing, synchronized lane-changing, and cooperative lane-changing are distinguished. To calibrate and validate the model, the data from a segment of highways was applied. The results demonstrated the reproduction of reality in terms of lane volume distributions and lane-specific speeds. 

Due to the lack of flexibility under dynamic driving situations and the resulting poor performance, data-driven approaches are motivated by training properly on large sample datasets. For instance, a neural network \citep{ren_new_2019}, a deep belief network (DBN) \citep{xie_data-driven_2019}, and a support vector machine (SVN) \citep{liu_novel_2019} are applied to model lane-changing decisions. Additionally, deep reinforcement learning (DRL) also shows great potential \citep{wang_reinforcement_2018,shi_driving_2019,peng_integrated_2022}. Since a lane-changing process incorporates a sequence of actions and the action to be executed affects the ultimate goal of the task, RL shows great potential to deal with this kind of problem. However, the mapping from state-action pairs to the total return (usually called Q-value) increases significantly with the size of state-action spaces, thus neural networks are applied to model this mapping. 

\subsection{Cognitive models}
\begin{figure}[t] 
\includegraphics[width=0.45\textwidth]{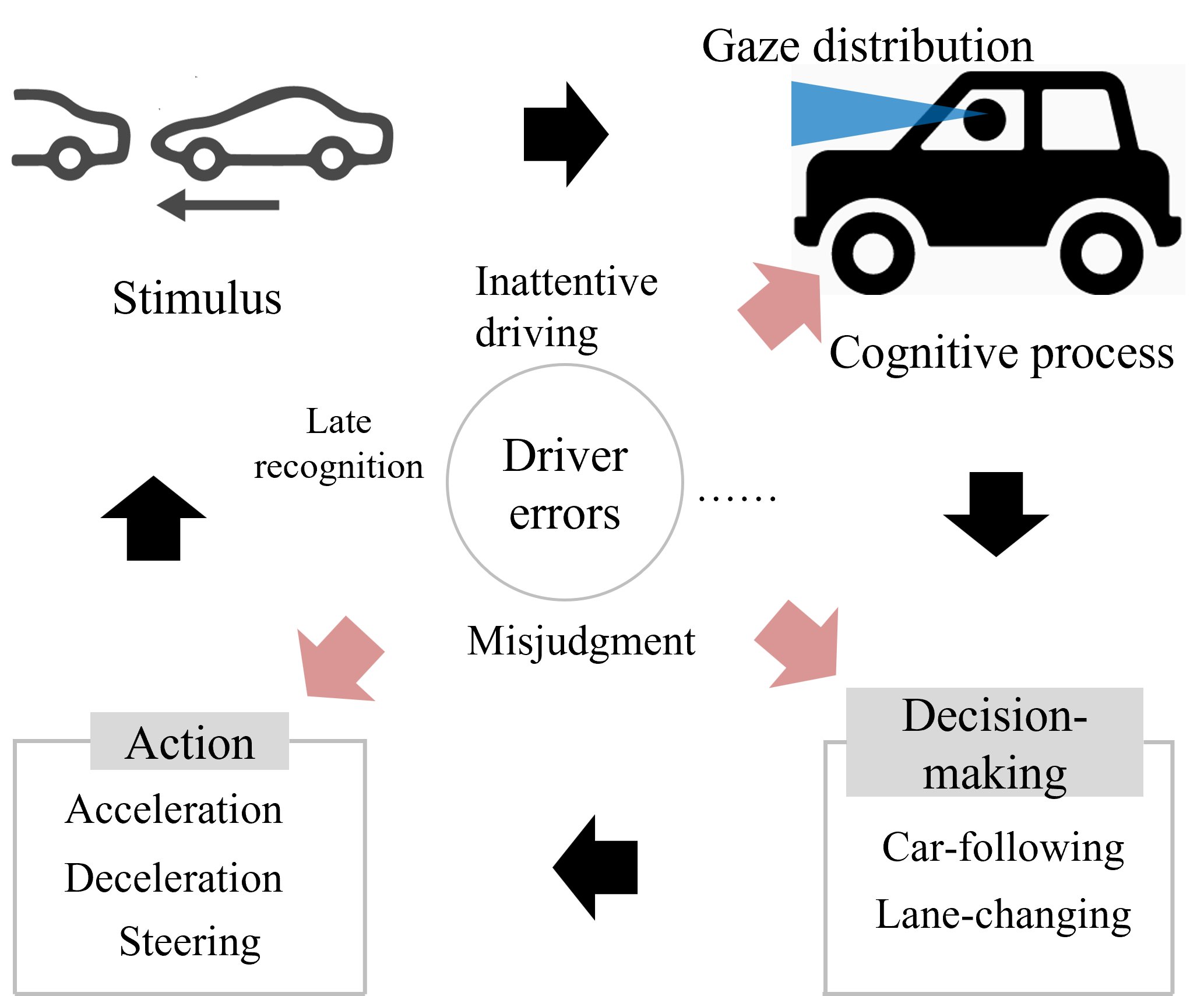}
\centering
\caption {\label{fig: cognitivemodel} The information flow to build a cognitive model. The driver perceives the stimulus and processes the information under possible driver errors. Based on the situation understanding result, decisions are made, where car-following and lane-changing models are applied. Finally, actions are executed in response to the stimulus. } 
\end{figure}

The goal of cognitive models is to simulate the human cognition process while driving, which includes perception, recognition, decision-making, and action, as illustrated in \cref{fig: cognitivemodel}. As the human cognitive process is considered, driver errors such as inattentive driving and misjudgment can occur in the perception, decision-making, and action processes. The modeling of driver errors is a distinct difference compared to those predictive models. Original cognitive models such as ACT-R \citep{anderson_integrated_2004}, Soar \citep{aasman_modelling_1995}, and QN-MHP \citep{liu_queueing_2006} are based on psychological cognitive architectures. These models can facilitate the understanding of driver behavior in the context of general human abilities and constraints. However, they are not suitable for simulation in arbitrary dynamic environments due to their complex structures.  

Recently, cognitive models aiming to simulate realistic traffic environments have been proposed. Driver failures that cause crashes can be simulated by taking into account inattentive or distracted driving during the information acquisition process in the cognitive models. A multi-agent traffic simulation software named Re:sim is proposed in \citep{kitajima_nationwide_2022} to model driver agents and their interactions with AVs. In the model, a driver agent perceives the surrounding objects in his line of sight and field of view. The relative states of the observed objects are then calculated, and recognition labels such as preceding or oncoming are assigned to them. Subsequently, hazardous objects are identified, and the risk of collision is estimated. Based on this judgment, the agent decides to operate and react using driving models such as the Weidemann following model \citep{wiedemann_simulation_1974}.

Similarly, the stochastic cognitive model (SCM) \citep{witt_modelling_2018,fries_driver_2022} consists of six modules: information acquisition, mental model, situation manager, action manager, action implementation, and driver characteristics. In information acquisition, the visual perception of the driver for perceiving the environment, such as the gaze allocation and fixation duration on a specific area of interest, is modeled. The mental model calculates and stores relevant driving states of observed objects. The situational risk is evaluated by the situation manager, which controls the action manager to provide an appropriate action for the action implementation. Importantly, the SCM model provides the opportunity to parameterize the driver characteristics so that the driver's perception and cognition, compliance with traffic rules can be flexibly adjusted. Unlike the SCM model, the DReaM model \citep{siebke_report_2021} focuses on urban traffic, particularly junction scenarios. Aside from that, the DReaM model has a similar structure to the SCM model.

\section{Driver models as references} \label{Driver models as references}
Driver models used as references for AV safety verification shall represent careful and competent drivers' driving abilities. A less skilled and experienced driver doesn't qualify to be used as a reference. Otherwise, the safety performance of an AV could be unsatisfying, if a bad reference is referred. To develop careful and competent driver models, critical scenarios are generally essential since excellent driving abilities are typically shown in critical scenarios. Note, once a careful and competent driver model is available, it can be used as a reference in both nominal and critical situations. 
As discussed in \cref{requirements}, risk-aware and capability dominate driver models are needed. More specifically, a driver is capable of identifying a risk in time to avoid an unnecessary intense response to a situation, and meantime all possible maneuvers to avoid or mitigate a collision are utilized.

Thus, we discuss in the following how the different driver models consider risk awareness. Braking is the most common reaction of drivers in critical situations. However, research \citep{eidehall_toward_2007} shows that if a collision can not be avoided by braking only, steering behavior is also performed by drivers. The combination of braking and steering has the potential to reduce the probability of a collision further. Therefore, all these three collision avoidance maneuvers are studied to present a holistic overview of driver models in critical situations. Cognitive models are not discussed even though some of them such as SCM \citep{fries_driver_2022} are supposed to be applicable in critical situations, their validity is not demonstrated. 

\subsection{Braking models} \label{brakingmodels}
We raise two questions during the review process to guide us in selecting driver models to fulfill the "risk-aware" and "capability dominant" requirements.
\begin{itemize}
  \item \textbf{Q1}: \textit{What conditions cause an emergency braking maneuver to be activated?} 
  \item \textbf{Q2}: \textit{What models are appropriate for describing emergency braking maneuvers?} 
\end{itemize} 

Regarding the triggering strategy \textbf{Q1}:
visual perception and critical metrics are usually used. Visual looming is a typical representative of visual perception, which refers to the optical size and expansion of a preceding vehicle on the retina  \citep{fajen_calibration_2005,markkula_farewell_2016}. To quantify this visual looming, the inverse tau \citep{lee_theory_1976}, defined as $\tau^{-1}= \dot{\theta} / \theta$, is applied, where $\dot{\theta}$ represents the preceding vehicle's optical expansion rate on the driver's retina, and $\theta$ is the optical size. The inverse tau increases with the collision risk level. However, Markkula \citep{markkula_modeling_2014} argued that a driver's braking is not initiated by exceeding a perceptual threshold, but by the accumulation of noisy perceptual evidence over time. Following this idea, Svärd et al. \citep{svard_computational_2021} proposed a driver model for initiating and modulating pre-crash brake response to deal with off-road glance behavior. In this model, the initiation time is obtained by the noisy accumulation of perceptual evidence for and against braking. Similarly, Waymo proposes a surprise-based stimulus, which is determined by the violations of the subject's current belief to its prior belief \citep{engstrom_modeling_2022}.

Besides visual perception, criticality metrics are employed to estimate situation risk. In \citep{un_ece_regulation_2021}, hard braking is applied when a challenging vehicle cuts in and the Time-to-collision (TTC) is smaller than 2 s. According to the study in \citep{schneemann_analyzing_2016}, the TTC for braking onset in urban environments is between 3 and 4 s, while the threshold for participants in the driving test is 2.5 s. Besides TTC, time-to-brake (TTB) is also used in some cases, which means the time left to avoid a critical situation by braking. For instance, a value of 0 is used as the threshold to activate the emergency braking maneuver in \citep{keller_active_2011}. Some driver models aim to model risk without an explicit threshold. Depending on the risk level, different deceleration values are applied. The fuzzy safety model (FSM) \citep{mattas_fuzzy_2020} models the longitudinal risk by defining a safe and an unsafe distance. If the actual distance is larger than the safe distance, no risk exists. Conversely, the highest risk is shown if the actual distance is below the unsafe distance. The risk is interpolated if the distance is between the two distance boundaries. Similarly, the risk-response driver model \citep{zhao_how_2020} utilizes risk field theory to model situation risk and then respond correspondingly according to the risk level. 

For emergency braking model \textbf{Q2}:
The car-following models discussed in \cref{car_following_models} are inappropriate to be used as a comparison reference for AV safety verification because they do not focus on a driver's braking reaction process to imminent situations, but rather on kinematic behavior at the vehicle level in nominal situations. The study in \citep{markkula_review_2012} shows that the Gipps and GHR models exhibit unsatisfying behavior in critical scenarios. Depending on the modeling method, emergency braking models to answer \textbf{Q2} can be roughly divided into two categories: "last second" braking models and risk anticipated braking models.

\textit{"Last second" braking models:} 
These models are designed to simulate the response delay that divers may experience during emergency braking. The Japanese driver model proposed in \citep{experts_of_japan_competent_2020} is a typical one in this category, where only braking is considered for collision avoidance. The driver model is separated into three segments: "Perception", "Decision" and "Braking". A risk perception point is defined to activate the decision and braking. In cut-in scenarios, the risk begins when the cut-in vehicle exceeds the normal lateral wandering zone and the TTC is below 2 s. Once risk is identified, the driver releases the gas pedal and steps on the braking pedal, this reaction delay time is 0.75 s. The deceleration rate then increases linearly until it reaches the max deceleration of 0.744 g. Subsequently, the maximum deceleration rate is maintained. 

%Meantime, the Reg157 model is defined in \citep{un_ece_regulation_2021}. This model utilizes TTC to estimate situation risk and apply braking maneuvers to avoid collisions. The maximum deceleration is assumed to be at least 6$\mathrm{~m} / \mathrm{s}^{2}$, and the perception time, together with the time needed to achieve the maximum deceleration, is equal to 0.35 s. 

The Responsibility Sensitive Safety model (RSS)\citep{shalev-shwartz_formal_2017} describes the rules that an AV should follow in order not to cause accidents proactively. The longitudinal safety distance considers the worst situation where the preceding vehicle decelerates with maximum deceleration, and the following vehicle accelerates with maximum acceleration and then decelerates moderately after the reaction time. A conservative distance is obviously assumed from this definition. Similar to RSS, the Safety Force Field \citep{nister2019safety} aims to guarantee that no unsafe situations are caused by the ego. If all actors in traffic had such a guarantee, no unsafe situations or collisions would occur. However, if the penetration rate of AVs is quite low, AVs will show very conservative behavior and may not be acceptable if the RSS model is utilized as the safety requirement for AVs. This unpleasant driving behavior will be lessened as the penetration rate increases. 

Since the driver reaction time is an important parameter in this type of model and varies among drivers and situations, a classification is made based on the characteristics of reaction time: fixed reaction time \citep{bando_analysis_1998,treiber_delays_2006}, variable buffer-based reaction time \citep{basak_modeling_2013,witt_modelling_2018} and random sampling reaction time \citep{przybyla_simplified_2012}. Fixed reaction time means that the reaction time is a fixed value. Variable buffer-based reaction time enables the selection of different independent variables (e.g., speed, distance, or indicator state). For each selected variable, a reaction time is drawn based on the underlying distribution. The random sampling of brake reaction time attempts to model distraction by using sufficient statistical samples to characterize stochastic distributions of reaction time, which can be fed into a braking model for crash prediction \citep{przybyla_simplified_2012}.  
Despite its simplicity, the reaction time model is assumed to be an important role in critical scenarios, describing the driver's extreme operating behavior in emergency situations.

\textit{Risk anticipated braking models:} 
Different from the "last second" braking models, the deceleration profile is adjusted according to the risk level in risk anticipated reaction models. Warren \citep{warren_dynamics_2006} described the braking process based on the tau theory, where the brake-pedal position $z$ is adjusted according to \cref{eq:TauModel}. $b_\mathrm{pedal}$ is a stiffness parameter determining the speed of pedal adjustments. $\varepsilon$ is the noisy term. $\dot \tau_m$ is the target margin value and $\dot \tau$ is the change rate of $\tau$.
\begin{equation} \label{eq:TauModel}
\Delta z = b_\mathrm{pedal}(\dot \tau_m - \dot \tau) + \varepsilon
\end{equation}

Similar to the $\tau$ model, the deceleration-error model \citep{fajen_calibration_2005} adjusts the deceleration by comparing the current deceleration to an ideal deceleration, where the ideal deceleration is defined as the deceleration when $\dot \tau$ equals -0.5. To capture the flexibility of the stiffness parameter in the deceleration-error model, an action boundary for describing the braking urgency is defined in \citep{harrison_affordance-based_2016}, beyond which a collision is unavoidable. If a situation becomes urgent, the proximity to the action boundary decreases, and the driver should apply braking with increasing strength.

Braking models using fuzzy logic theory show a similar concept, where human decision-making processes are approximated \citep{zhang2011user} according to the perceived risk. In addition, the possibility of considering various driving styles and driving environments \citep{fernandez2016driver} can be included in fuzzy logic. Three steps are included in the fuzzy reasoning process. Fuzzification converts the input values to fuzzy values based on predefined rules. Then, the inference engine mimics human reasoning by performing fuzzy inference on the inputs. The output fuzzy variables are finally converted to executable values by defuzzification. The states of the following vehicle and the preceding vehicle are commonly used as input values, and the output values are brake angle or brake pressure \citep{mamat_fuzzy_2009,basjaruddin_hardware_2016,rizianiza_automatic_2021}. 
\begin{figure}[!ht]
\includegraphics[width=0.48\textwidth]{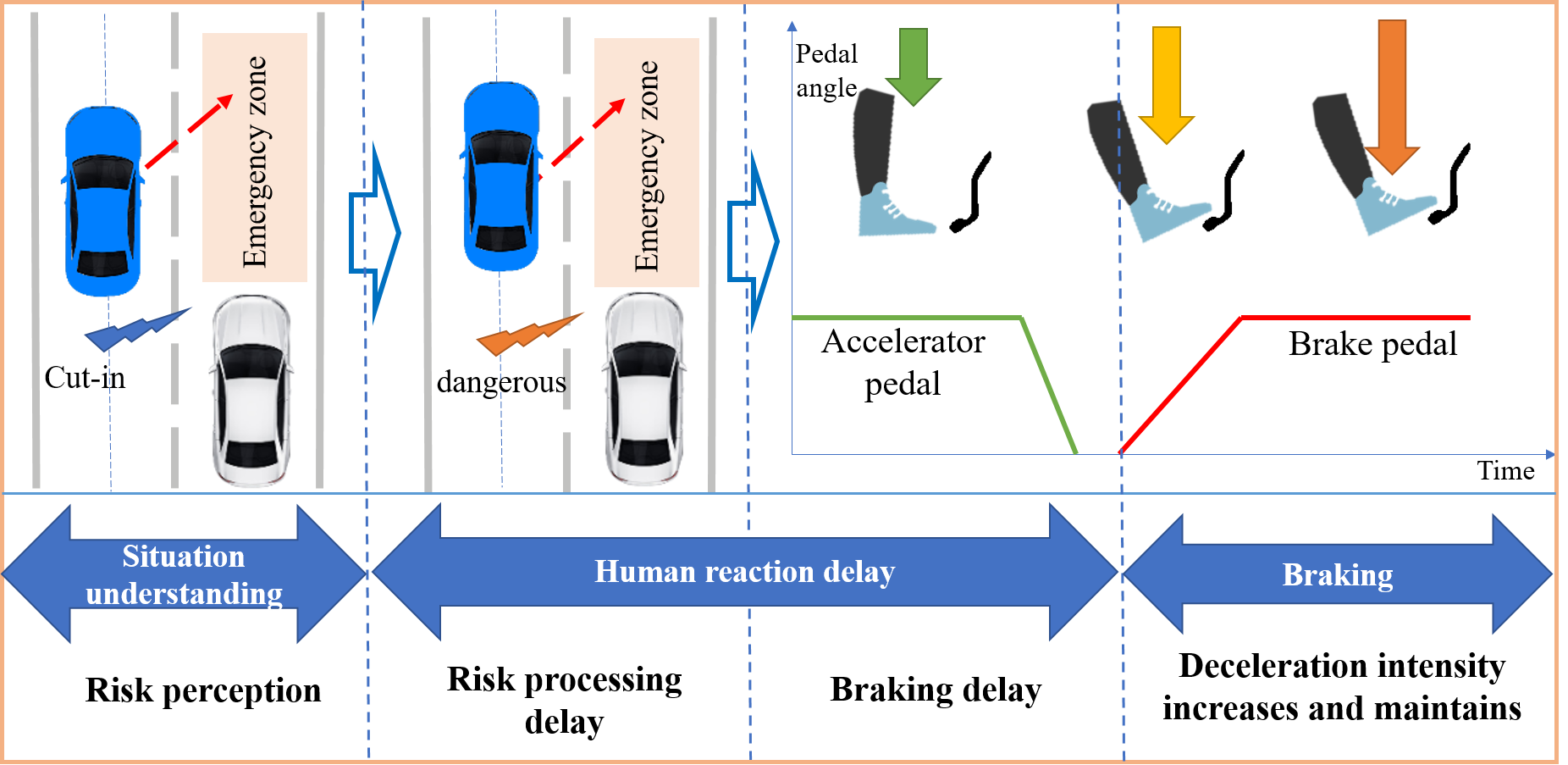}
\centering
\caption {\label{fig: ReacitonTimeBraking} An illustration of the braking process for collision avoidance. The situation risk is estimated by a risk perception metric. When an emergency braking maneuver is necessary, deceleration onsets after a certain reaction time. The maximum deceleration is either constant or adjusted according to the situation risk level.} 
\end{figure}

Recently, the Fuzzy Safety Model (FSM) \citep{mattas_driver_2022} for rear-end collisions is proposed. Depending on the scenario type, longitudinal and eventually lateral distances are checked against the safe distance to judge if the braking should be initiated. Unlike previous studies, two fuzzy surrogate safety metrics are explicitly employed to evaluate the situation risk. Subsequently, a proper deceleration corresponding to the risk level is performed.

\begin{comment}    
according to \cref{eq:FSM}.
\begin{equation} \label{eq:FSM}
    D =     
    \begin{cases}
      car-followingS(D_\mathrm{max} - D_\mathrm{comf}) + D_\mathrm{comf} & \text{if $car-followingS > 0$}\\
      PFS \cdot D_\mathrm{comf} & \text{if $car-followingS = 0$}\\
    \end{cases} 
\end{equation}
where the Proactive Fuzzy surrogate Safety metric (PFS) and the Critical Fuzzy surrogate Safety metric (car-following) are the two metrics to evaluate situation risk. $D_\mathrm{max}$ and $D_\mathrm{comf}$ represent the maximum and comfortable deceleration of the following vehicle.
\end{comment}

From the above analysis, we discover the following commonalities of the discussed collision avoidance braking models. First, a risk perception metric is essential in order to trigger the braking model. Either visual perception or criticality metrics are applied to estimate the situation risk. Second, there is a distinction between "last second" and "risk anticipate" braking models. Models like RSS have fixed deceleration profiles, while other models like FSM have adjustable deceleration profiles depending on the risk level. We use \cref{fig: ReacitonTimeBraking} to summarize the braking models. Due to the special application for AV safety verification, interpretability is an important factor for the models in order to provide an understandable reference for such safety-critical systems. Additionally, near-crash and crash data are limited, hindering the application of data-driven models in this case.

\begin{comment}
\begin{figure}[!ht]
  \begin{minipage}[t]{0.95\linewidth}
    \includegraphics[width=\linewidth]{photo/steering1.png}%
  \end{minipage}\hfil
  
  \begin{minipage}[t]{0.95\linewidth}
    \includegraphics[width=\linewidth]{photo/steering2.png}%
  \end{minipage}\hfil
  
  \begin{minipage}[t]{0.95\linewidth}
    \includegraphics[width=\linewidth]{photo/steering3.png}%
  \end{minipage}%
  \caption {\label{fig:steering_model} The illustration of three evasive steering models using the preview method. (a) models using a single preview point; (b) models optimizing over a preview horizon; (c) models using multiple preview points.} 
\end{figure}
\end{comment}

\subsection{Evasive steering models} \label{evasivesteeringmodel}
Similar to the braking models, two research questions are proposed based on the "risk-aware" and "capability dominant" requirements:
\begin{itemize}
  \item \textbf{Q1}: \textit{What conditions cause an evasive steering model to be activated?} 
  \item \textbf{Q2}: \textit{What models are appropriate for describing evasive steering maneuvers?} 
\end{itemize}

For the triggering strategy \textbf{Q1}:
Time-to-Collision (TTC) is the most common metric to activate evasive steering maneuvers despite its simplicity. For instance, A TTC threshold of 2.5 s is utilized in \citep{llorca_autonomous_2011}. They demonstrated that the proposed approach can perform human-like pedestrian collision avoidance maneuvers by steering with a maximum driving speed of 30 km/h. Zhao et al. \citep{zhao_emergency_2019} consider 0.75 s for the threshold of TTC for an emergency evasion maneuver. Similarly, a two-dimensional TTC with a threshold of 5 s is defined for a data-driven evasive steering model \citep{guo_modeling_2022}, while Time-to-Steer (TTS) with a value of 0 is used \citep{keller_active_2011} to initiate an evasive maneuver. TTS represents the time left to avoid a critical situation by steering. 

A "threat metric" related to acceleration or jerk level is utilized to trigger an evasion steering maneuver, which is defined by a cubic polynomial with zero derivatives at the knots. The number of knots and their position depends on the number of objects along the path and their position. However, the threshold of the "threat metric" is not presented \citep{eidehall_real_2013}.

Some studies define the trigger moment implicitly. Shiller \citep{shiller_emergency_1998} suggested using the clearance curve to determine the "last point" for an evasive steering maneuver, where the "last point" was the minimal longitudinal distance beyond which an obstacle cannot be avoided by steering at a given initial speed. Similarly, Isermann et al. \citep{isermann_collision-avoidance_2012} argue that the timing to evade is determined by at what distance the evasion must be executed so that a collision is still preventable. They utilize a sigmoid function to describe the evasive trajectory. Park et al. \citep{park_emergency_2021} assume that braking is first applied, and the steering maneuver must wait until collision by braking is no longer possible. Furthermore, the required lateral acceleration should be larger than a threshold beyond which the collision can be averted by steering. 

For the evasive steering model \textbf{Q2}: We divide the existing evasive steering models into three categories based on the various ways to generate the evasive trajectories: analytical trajectories, optimization-based, and data-driven models. 
Note we do not focus on the execution of the evasive maneuver, which is the task of a controller. Instead, generating an evasive trajectory is the topic once an evasive maneuver is essential.

Mathematical expressed analytical trajectories are commonly used to generate evasive steering maneuvers. For example, sinusoidal and exponential curves are applied to generate evasive lateral maneuvers with a given lateral offset \citep{durali_collision_2006}. In addition, a fifth-order polynomial in curvilinear coordinates is employed for the purpose of minimizing jerk in \citep{dona_stability_2019}. Since a fifth-order polynomial has a zero lateral velocity and acceleration at both the initial ($X_0$, $Y_0$) and the final position ($X_e$, $Y_e$), it ensures the vehicle's stability after avoidance. An implementation of a fifth-order polynomial is demonstrated in \citep{soudbakhsh2013vehicle, he_emergency_2019}, as expressed in \cref{polynomial}.

\begin{equation} \label{polynomial}
Y(X) = Y_e[10(\frac{X}{X_e})^3 - 15(\frac{X}{X_e})^4 + 6(\frac{X}{X_e})^5]
\end{equation}

In \citep{soudbakhsh2011emergency, soudbakhsh2015steering}, a trapezoidal acceleration profile is utilized to generate the evasive trajectory. Based on this concept, Dona et al. \citep{dona_towards_2023} study a simple kinematic evasive steering model to the end of estimating the minimum time to perform an evasive steering maneuver. They argue that simplicity and generality are important for regulatory adoption. To the authors' best knowledge, this is the first work that attempts to derive a safety benchmark model for evasive steering maneuvers until now. However, whether the model is capability dominant is unclear. A comparison between human drivers' evasive action time \citep{shah_analysis_2021} and the minimum time derived from the model is meaningful.

Apart from using predefined analytical trajectories, an alternative is the optimization-based trajectory generation. Typically, an objective with some constraints is defined. The problem is then solved by optimization techniques such as IPOPT solver \citep{wachter2006implementation}. In \citep{soudbakhsh2013vehicle}, lateral acceleration and steering rate are defined as the optimization objectives. They show that the proposed trajectory optimization technique requires less time than that of the trapezoidal acceleration profile for an evasive steering maneuver. The model predictive control (MPC) is also widely used for evasive trajectory generation \citep{park_emergency_2021, wurts2021collision} because of the advantages of an explicit definition of constraints and sub-optimal performance \citep{shim2012autonomous}. However, the vehicle dynamic model and even the tire model considered in MPC may hinder its application as a reference because vehicle-specific calibration is needed.

Lastly, A deep deterministic policy gradient (DDPG) algorithm \citep{guo_modeling_2022, feher2020hierarchical}, which could learn the sequential decision-making process over continuous action spaces, was used to model evasive behaviors. Another data-driven model proposed by Das and Mishra \citep{das_machine_2022} attempted to avoid collisions using left and right turn maneuvers that are learned from a dataset. We found that few machine learning-based evasive steering models are currently available. This may be due to the lack of training data in critical situations. 

\begin{figure}[!ht]
\includegraphics[width=0.5\textwidth]{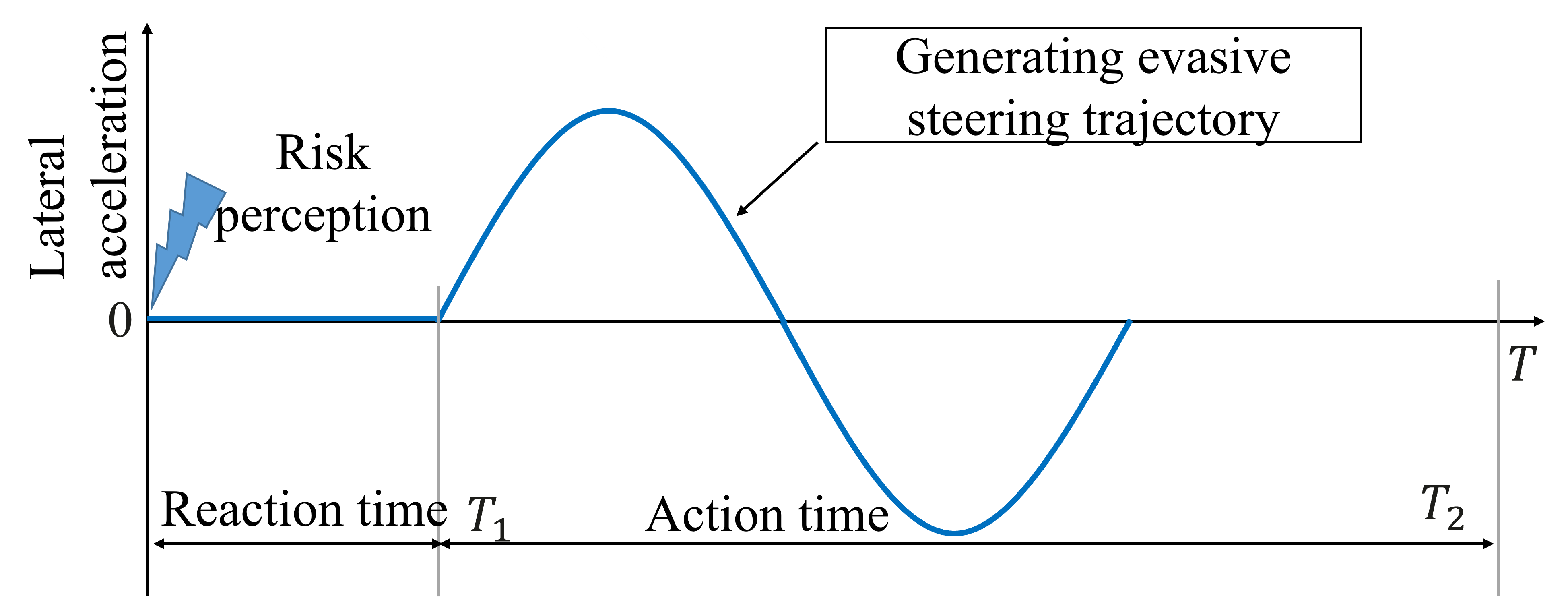}
\centering
\caption {\label{fig: SteeringIntroduction} The process of an evasive steering maneuver performed by a driver. If a driver perceives a situation risk that necessitates a response, the steering maneuver is executed by changing the lateral acceleration after the reaction time.} 
\end{figure}

From the above analysis, we could find that most literature focuses on the modeling of evasive maneuver itself, while limited research is conducted about human drivers' evasive capabilities in critical situations \citep{zhao2018emergency,shah_analysis_2021}. As a result, it is difficult to argue the proposed reference models are capability dominant. Additionally, analytical trajectories are promising because of their simplicity and less reliance on specific vehicles. Finally, we use \cref{fig: SteeringIntroduction} to illustrate the general process of the evasive maneuver in case of a critical situation. The situation risk is estimated. When the evasive maneuver is possible, e.g., enough free space at the steering side, the lateral acceleration changes accordingly to finally generate the evasive trajectory after a certain reaction time.

\begin{table*}[!ht]
\caption {\label{tab:tableII} The metrics for evaluating driver models in terms of their applications in AV safety assessment } 
\begin{tabularx}{\linewidth}{p{1.3cm} p{1.8cm} p{7.5cm} Z Z}
\toprule
Aims & 
\makecell[l]{Evaluation \\ metrics} & 
\makecell[l]{Descriptions} & 
\makecell[l]{Typical models} &
\makecell[l]{Possible \\ applications} \\ 
\midrule
   \multirow{28}{*}{Simulation} 
   & Accuracy & Accuracy refers to the degree to which a driver model is able to reproduce the actual driving behavior of human drivers considering the "interaction-aware" and "human-like" requirements. An accurate driver model will produce results that are very close to the actual human driving behavior in different scenarios. This is important for a valid and credible AV safety assessment.  & Machine learning-based driver models \citep{xie_data-driven_2019,wang_capturing_2018} & Microscopic simulations \\  \cline{2-5}
   & Adaptability & Driver models shall demonstrate their adaptability to different situations. Human driver behavior can vary across different situations. Thus, the driver model needs to adapt to these varying situations and reflect the corresponding changes in driver behavior. This adaptability is essential for accurately modeling human driving behavior and evaluating AV safety performance in simulations. & The models considering driver characteristics \citep{liao_car-following_2019,zhang_study_2018} & Mileage-based simulation \\  \cline{2-5}
   & Variability & Driver models shall reflect the variability in human driver behavior, which arises from the complex processes of perception, cognition, and decision-making. Even for an identical driver in the same situations, various behaviors including errors may be exhibited. Those driving behavior uncertainties are valuable to test the robustness and safety of AVs in simulations. & The model with stochastic brake reaction time \citep{przybyla_simplified_2012}; the SCM \citep{fries_driver_2022}  & Testing AVs' ability to handle the behavioral diversity of surrounding human drivers within the same scenario. \\ \cline{1-5} 
   \multirow{10}{*}{Reference} 
   & \makecell[l]{Risk awareness} & The driver model shall perceive risk in time to present "careful". Careless and conservative drivers are not suitable to be treated as benchmarks for AVs.  & The FSM \citep{mattas_fuzzy_2020} & Deriving safety requirements for AVs. \\ \cline{2-5} 
   & Capability domination & The driver model shall be representative of human drivers to present "competent". A less skilled or experienced drivers are not suitable to be treated as benchmarks for AVs. & The Japanese driver model \citep{un_ece_regulation_2021} & Deriving safety requirements for AVs. \\ \cline{2-5} 
   & Interpretability & Using a driver model as a benchmark for evaluating AV safety performance means that it serves as a criterion in the evaluation. As a criterion, its definition should be clear and interpretable to ensure the reliability of the evaluation and gain the public's trust. & The RSS model \citep{shalev-shwartz_formal_2017}; the analytical models \citep{he_emergency_2019} & Defining benchmark for AV safety assessment \\ 
\bottomrule
\end{tabularx}
\end{table*}

\subsection{Braking and steering models}
To avoid collisions, braking is an effective measure. The evasive steering maneuver allows a later intervention than the braking maneuver, while simultaneously braking and steering maneuvers can further reduce the evasive length and enable a further later intervention \citep{ackermann2015collision}. Thus, simultaneously braking and steering are considered the very "last second" action \citep{choi2017simultaneous}. The driver model proposed by Jurecki and Stańczyk \citep{jurecki_driver_2009} synthetics these two maneuvers analytically. In the driver model, the braking model is described as follows:
\begin{equation} \label{eq:SteeringandBraking1}
\textit{D} + \textit{W}_1 \dot{D} = \textit{W}_2\Delta{y}(t - T_\text{R}) + \textit{W}_3 \frac{v_{\text{rel}}}{d_{\text{rel}}}
\end{equation}
where \textit{D} is deceleration; $\Delta{y}$ represents the lateral relative distance; $T_\text{R}$ is the driver's reaction time; $\textit{W}_1$, $\textit{W}_2$ and $\textit{W}_3$ are model constant parameters. $v_{\text{rel}}$ and $d_{\text{rel}}$ are the longitudinal relative velocity and distance, respectively. The corresponding steering model $\delta$ is expressed by:
\begin{equation} \label{eq:SteeringandBraking2}
\delta + \textit{W}_4 \dot{\delta} = \textit{W}_5 \Delta{y} (t - T_\text{R})
\end{equation}
$\textit{W}_4$ and $\textit{W}_5$ are another model parameters. These model parameters are identified based on 450 trials with 30 drivers.

Schorn and Isermann \citep{schorn_automatic_2006} employ a sigmoidal function to generate the desired trajectory, which is followed by a feedforward control to execute braking and steering. Similarly, many studies \citep{falcone_model_2007,hajiloo_integrated_2021,li_emergency_2022,wang_integrated_2022} nowadays use the model predictive control (MPC) technique to control the steering angle and the brakes with given constraints and objectives. As a result, the optimization-based methods demonstrate a tendency to deal with simultaneous braking and steering. 

Recently, Waymo developed a non-impaired with eyes on the conflict (NIEON) reference driver model \citep{scanlon_collision_2022}, with the goal to provide a benchmark for AV safety assessment. For braking actions, NIEON applies a constant jerk up to a maximum deceleration, which is similar to the Japanese driver model. The difference is that they employ surprise-based response timing defined by Waymo as well \citep{engstrom_modeling_2022}. For the evasive steering model, they used a pure pursuit controller to follow the pre-maneuver trajectory at a given preview point. They demonstrated that the NIEON can be used as a benchmark for AVs. The model is a pioneering reference driver model that considers both braking and evasive steering maneuvers although the generation of pre-maneuver trajectory is not fully described.

\section{Applicability} \label{Applicability}
In this section, we first define evaluation metrics for driver models in the two applications discussed in the paper. Next, we summarize and categorize the aforementioned driver models based on their model characteristics. We then analyze the potential applications and suitability of different driver model categories for the two different applications based on the proposed evaluation metrics to answer the \textbf{RQ3} finally.

\subsection{Evaluation metrics}
As we assume a low penetration rate of AVs, the surrounding vehicles are controlled by driver models in simulations. To simulate realistic traffic scenarios for testing AVs and increase the credibility of the testing results, the driver model shall meet the requirements discussed in \cref{requirements}. Thus, we define three concrete evaluation metrics: accuracy, adaptability, and variability, which are presented in \cref{tab:tableII}, along with their descriptions and examples of typical models that best meet each evaluation metric. Additionally, possible applications are provided, for which this evaluation metric is particularly relevant. 

A reference driver model, on the other hand, used as a benchmark for AVs, shall represent the capabilities of careful and competent drivers. Its underlying logic is that an AV shall demonstrate superior performance compared to a careful and competent human driver in the same testing scenarios, i.e., it can successfully avoid collisions in those scenarios in which a careful and competent human driver can do. Considering the requirements derived in \cref{requirements}, a reference driver model for AV safety evaluation shall satisfy three evaluation metrics: risk awareness, capability domination, and interpretability, as depicted in \cref{tab:tableII}. As a benchmark for evaluating the performance of AV, interpretability and transparency are crucial for regulation \citep{fagnant_preparing_2015} and insurances \citep{grosso_how_2021}.

Aside from the evaluation criteria provided in \cref{tab:tableII}, we additionally consider the simplicity of each sort of driver model, which is desirable though not mandatory. To capture the diverse human driver behavior and different driving situations, simulations must be run on many scenarios or driving miles, often necessitating the use of methods such as the Monte Carlo method. Moreover, concrete scenarios frequently involve multiple vehicles controlled by human drivers. Therefore, reducing the complexity of the driver model can help to minimize computational costs.

\subsection{Applicability analysis}
This subsection evaluates the summarized driver models in terms of their applications for AV safety assessment. 
First, we discuss the driver models in terms of their suitability for simulation-based testing. We divide the driver models into the following four categories. \textit{Linear model}: a linear relationship between the factors considered and the output of the driver model (e.g. acceleration) is modeled, such as the GHR \citep{xie_data-driven_2019} car-following model; \textit{Non-linear model}: the relationship between the stimulus and the output of the model is nonlinear or probability distributed, such as fuzzy theory-based model \citep{kikuchi_car-following_1992} and stochastic reaction time model \citep{przybyla_simplified_2012}; \textit{Data-driven model}: data-driven models capture the interaction and driving behavior by training on data, such as the recurrent neural network in \citep{zhou_recurrent_2017} and deep bayesian network model in \citep{xie_data-driven_2019}.

%Based on the proposed metrics, a comparison of linear, non-linear, optimization-based, and data-driven models is conducted. As a result, the dis- and advantages of each type of driver model are presented, which can guide developers to select proper driver models for their different purposes.  

\cref{tab:tableIII} compares the above-categorized driver models in meeting the proposed evaluation metrics for simulations. In the table, "+" symbol indicates that the model is able to meet the evaluation metric. Conversely, "-" means the metric is challenging for the model. Nonlinear and data-driven models demonstrate superior capability in meeting more evaluation metrics than linear models, particularly in three out of the four evaluation metrics. The key distinguishing factor between nonlinear and data-driven models is the trade-off between complexity and accuracy. Nonlinear models exhibit relatively lower complexity and are thus more appropriate for scenarios that demand extensive simulation calculations, such as those involving high mileage and multi-scenario coverage. In contrast, data-driven models, particularly those utilizing deep learning techniques, consider more feature dimensions and exhibit higher-order nonlinearity, which leads to higher accuracy but also entails increased complexity and computational costs.

\begin{table}[!ht]
\caption {\label{tab:tableIII} Evaluation and Comparison of Driver Models with Different Model Characteristics in Terms of their suitability for Simulation Purposes} 
\begin{tabularx}{\linewidth}{c c c c}
\toprule
Evaluation metrics & 
\makecell[c]{Linear models} & 
\makecell[c]{Non-linear \\ models} &
\makecell[c]{Data-\\ driven} \\ 
\midrule
    Variability & - & +  & + \\
    Adaptability & - & +  & + \\
    Simplicity & + & +  & - \\
    Accuracy & - & -  & + \\
\bottomrule
\end{tabularx}
\end{table}

%In the application where driver models serve as a reference, to meet the evaluation metric of maneuver- coverage, the ideal model should choose braking, steering, or a combination of both based on the current situation, just like a CC driver. As discussed in \cref{Driver models as references}, only a few models consider collision avoidance through simultaneous braking and steering. Consequently, car-following and lane-changing models discussed above are typically needed to be combined to model diverse maneuvers in case of a critical situation, and a corresponding maneuver switch model is required. This section mainly evaluates individual categories, and the combination of different driver models is not considered in each category. Therefore, the metric of maneuver coverage is in the analysis here. In addition, regarding the evaluation metrics of representativeness and accuracy, we will also analyze the two maneuvers of braking and steering separately. 

In the application where driver models serve as references, the comparative results are summarized in \cref{tab:tableIV} and \cref{tab:tableV}. As described in \cref{brakingmodels}, we compare "last second" and risk anticipated braking models in terms of their application as benchmarks. while analytic models, optimization-based models, and data-driven models are employed for evasive steering models, as described in \cref{evasivesteeringmodel}.

The "last second" braking models such as \citep{shalev-shwartz_formal_2017,experts_of_japan_competent_2020} have the advantages of capability domination by good calibration on the data of trained drivers in critical braking experiments \citep{liu2021calibration}. Meantime, they generally have better interpretability and lower complexity because of explicitly defined math relations. However, they usually fail to meet risk awareness requirements because the activation time is fixed and usually very late to avoid false positive emergency braking. The risk anticipated braking models such as \citep{mattas_driver_2022}, on the other hand, are risk-aware and capability dominant. They also meet the requirements of interpretability and simplicity. Thus, they are preferred to "last second" braking models when deploying them as benchmarks for AVs. 

In lateral evasive situations, optimization-based and data-driven models are risk-aware since they consider safety in constraints or training data, while analytical models do not model safety inherently. In addition, analytical and optimization-based models are currently little studied, and few experiments demonstrate that they can represent the "competent" aspect. The data-driven models are supposed to reach the driving performance level of a skilled driver, as demonstrated in \citep{wu_end--end_2019}. Nonetheless, the analytic models show advantages in interpretability and simplicity, while the optimization-based models are generally understandable through explicitly defined objectives and constraints, but their simplicity is less satisfying due to the complex solving process. Because the studies of reference driver models for simultaneously braking and steering are few and currently limited to MPC techniques, a comparison with other models in terms of their suitability as benchmarks is not considered in the paper.

\begin{table}[!ht]
\caption {\label{tab:tableIV} Evaluation and comparison of braking models with different model characteristics in terms of their suitability for reference purposes} 
\begin{tabularx}{\linewidth}{c c c }
\toprule
\makecell[c]{Evaluation metrics \\ (Braking)} & 
\makecell[c]{"Last second" \\ braking model} & 
\makecell[c]{Risk anticipated \\ braking model} \\ 
\midrule
    Risk awareness & - & +   \\
    Capability domination & + & +   \\
    Interpretability & + & +  \\
    Simplicity & + & + \\
\bottomrule
\end{tabularx}
\end{table}

\begin{table}[!ht]
\caption {\label{tab:tableV} Evaluation and comparison of evasive steering models with different model characteristics in terms of their suitability for reference purposes} 
\begin{tabularx}{\linewidth}{c c c c}
\toprule
    \makecell[c]{Evaluation metrics \\ (Steering)} & 
    \makecell[c]{Analytical \\ models} & 
    \makecell[c]{Optimization- \\ based} & 
    \makecell[c]{Data- \\ driven} \\ 
\midrule
    Risk awareness & - & +  & + \\
    Capability domination & unkown & unkown  & + \\
    Interpretability & + & +  & - \\
    Simplicity & + & -  & - \\
\bottomrule
\end{tabularx}
\end{table}

\section{Discussion} \label{Discussion}
In this paper, we investigated the functions of driving models in the safety assessment of AVs. The guidance for safety engineers to select appropriate driving models for AV safety evaluation in simulations or for determining AV safety performance levels compared to careful and competent drivers was provided by addressing three research questions. Meanwhile, we compared the different driver models in terms of their suitability for simulations and references against proposed metrics and identified current gaps. To the authors' best knowledge, this is the first work that summarizes driver models in terms of their applications for AV safety assessment.

For \textbf{RQ1}, we first analyzed where driver models are used for AV safety assessment. We discovered that they are typically used in simulations to control surrounding vehicles for generating test environments of AVs and showcase new functions as benchmarks for AVs. Based on these potential applications, we discussed the corresponding requirements. More specifically, interaction-aware decision-making, human-like behavior, driver characteristics, and cognitive process are important for driver models in simulations, while risk-aware and capability dominate driver models are expected as AV safety evaluation references. These two different applications and the derived requirements pose a solid foundation for the driver models to be reviewed in the paper.

For \textbf{RQ2}, we presented an overview of driver models based on the defined scope in the paper. Various car-following and lane-changing models were described under the category of predictive models. It is observed that applying artificial intelligence (AI) to model both lateral and longitudinal driving behavior shows a trend to meet the diverse requirements. Due to the inscrutability of AI, Hybrid models that combine AI and analytic models to increase explainability while maintaining high fidelity are being investigated \citep{bhattacharyya_hybrid_2021}. Additionally, cognitive models that aim to model driving behavior in nominal traffic and critical scenarios show promising results except for the high model complexity. Regarding driver models as references, we outlined possible maneuvers, including collision avoidance by braking, collision avoidance by steering, and collision avoidance by simultaneous braking and steering to build capability dominate drivers. In addition, we considered the timing for triggering avoidance maneuvers, which is essential when initiating collision avoidance maneuvers.  

For \textbf{RQ3}, we first proposed metrics corresponding to the defined requirements to assess the reviewed driver models. Based on these metrics, we are able to analyze the strengths and weaknesses of each driver model, which facilitates the determination of its applicability. For the driver models for simulations, non-linear and data-driven models are promising. If higher accuracy is desired and the computational cost is acceptable, data-driven models are better choices. Conversely, non-linear models are preferred if a relatively lower accuracy can be tolerated in exchange for the lower computational cost. As computer hardware performance rapidly advances, the high complexity of data-driven models may no longer be a concern in the future, making them the optimal choice. Regarding driver models as references, analytical models are compelling since they can exhibit high interpretability and simplicity. However, whether they are capability dominant compared to skilled and experienced human drivers is less studied. Lastly, if the interpretability issue of data-driven models can be addressed in the future, they could also become a good choice.

Compared to the work in \citep{siebke_what_2022}, we analyzed in-depth various types of driver models as references for AV safety assessment and discussed their applicability by analyzing their strengths and weakness, while they focused on a discussion about human error modeling in traffic simulations. For some other surveys, either they only review one category of driver models like car-following models \citep{han_modeling_2022}, or driver models for AV safety assessment are not considered in their reviews \citep{singh_analyzing_2021,park_review_2022}. 

Based on the discussion in \cref{Functions and requirements}, We restricted our evaluation to driver models that are suitable for either AV safety assessment in a simulation environment or AV approval by displaying their benchmark functions. Thus, other topics like driver behavior analysis and error modeling are not considered. However, the relevant driving models are thoroughly identified. Thus, the discussion and the derived phenomena are pertinent. The proposed metrics to evaluate different types of driver models are summarized considering our two applications for AV safety assessment. Therefore, they are limited to our purpose, and a comprehensive evaluation of driver models, including their applications in other domains, should be further studied.    

Generally, our paper provides a fundamental consideration of currently existing driver models for AV safety assessment. On the one hand, it can aid policymakers, such as the working party of Harmonization of Vehicle Regulations in UNECE, to choose appropriate driver models when drafting regulations for AV approval. On the other hand, safety or simulation engineers could utilize suitable driver models depending on their demands to conduct their simulations to provide evidence to support a holistic safety argumentation for their developed AVs.

In summary, the paper surveys driver models to answer three research questions regarding their applications for AV safety assessment. By summarizing the identified relevant papers, we are able to present guidance on what type of driver models are suitable for what type of task. Compared to other related works, our work presents a holistic overview of driver models, focusing on their applications in AV safety assessment. The survey is generally helpful for policymakers and developers. 

\section{Conclusion and future work} \label{Conclusion and future work}
The paper presented a comprehensive review of existing driver models regarding their applications in AV safety assessment and discussed their suitability by proposing evaluation metrics derived from analyzed requirements. To the authors' best knowledge, this is the first work that summarizes driver models' functions for AV safety assessment. The work provides insightful guidance for policymakers and simulation engineers when formulating AV release regulations or conducting AV safety assessments in simulations.

Based on the results, we have the following findings: 1) AI-based driver models have attracted much attention recently to model both lateral and longitudinal driving behavior in nominal situations. In particular, reinforcement learning-based models in simulation-based testing are promising; 2) cognitive models show great benefit by including human errors in the models to simulate, for example, inattentive driving; 3) Analytical models has the advantages of interpretability and simplicity when used as benchmarks compared to optimization-based and data-driven models. Nevertheless, more studies about the capability of analytical models are needed to demonstrate their representativeness of competent human drivers.

Meantime, these findings indicate future working directions. More open-sourced driver simulator datasets and crash datasets are desirable in comparison to the number of open-sourced naturalistic driving datasets such as HighD \citep{krajewski_highd_2018}, AD4CHE \citep{zhang_ad4che_2023} and pNEUMA dataset \citep{barmpounakis2020new}. The data is valuable for studying driver steering, steering and braking behavior in critical situations. In this way, a holistic driver model incorporating different maneuvers at different stages of a collision could be created as a reference instead of a single emergency braking model. Explainable AI must be prioritized to achieve both high interpretability and performance. In all, the paper provides a solid foundation for future driver models for AV safety assessment.

\section*{Declarations of interest}
The authors declare that they have no known competing financial interests or personal relationships that could have appeared to influence the work reported in this paper.

\printcredits

\bibliographystyle{cas-model2-names}

\bibliography{references}

\end{document}